\documentclass[journal]{IEEEtran}
\usepackage{amssymb}
\usepackage[algoruled,vlined,linesnumbered]{algorithm2e}
\usepackage{comment}
\usepackage{amsthm}

\theoremstyle{definition}

\theoremstyle{remark}

\usepackage{bm}
\usepackage{cite} 
\usepackage{amsmath}
\usepackage{float}
\usepackage{graphicx}
\usepackage{subfigure}
\usepackage{xcolor}
\usepackage{array}
\usepackage{multirow}

\newcolumntype{C}[1]{>{\centering\let\newline\\\arraybackslash\hspace{0pt}}m{#1}}

\begin{document}
\title{Learning Deep Analysis Dictionaries {for Image Super-Resolution}}

\author{Jun-Jie~Huang,~\IEEEmembership{Member,~IEEE,} and~Pier
Luigi~Dragotti,~\IEEEmembership{Fellow,~IEEE}

\thanks{The authors are with the Department of Electrical and Electronic Engineering, Imperial College London, London SW7 2AZ, U.K. (e-mail: j.huang15@imperial.ac.uk).}

}

\maketitle

\begin{abstract}
Inspired by the recent success of deep neural networks and the recent efforts to develop multi-layer dictionary models, we propose a Deep Analysis dictionary Model (DeepAM) which is optimized to address a specific regression task known as single image super-resolution. Contrary to other multi-layer dictionary models, our architecture contains $L$ layers of analysis dictionary and soft-thresholding operators to gradually extract high-level features and a layer of synthesis dictionary which is designed to optimize the regression task at hand. In our approach, each analysis dictionary is partitioned into two sub-dictionaries: an Information Preserving Analysis Dictionary (IPAD) and a Clustering Analysis Dictionary (CAD). The IPAD together with the corresponding soft-thresholds is designed to pass the key information from the previous layer to the next layer, while the CAD together with the corresponding soft-thresholding operator is designed to produce a sparse feature representation of its input data that facilitates discrimination of key features. {DeepAM uses both supervised and unsupervised setup.}
Simulation results show that the proposed deep analysis dictionary model achieves 
{better performance compared to a deep neural network that has the same structure and is optimized using back-propagation when training datasets are small.}
{On noisy image super-resolution, DeepAM can be well adapted to unseen testing noise levels by rescaling the IPAD and CAD thresholds of the first layer.}
\end{abstract}


\IEEEpeerreviewmaketitle

\section{Introduction}

\IEEEPARstart{D}{eep} Neural Networks (DNNs) \cite{lecun2015deep}
are complex architectures composed of a cascade of multiple
linear and non-linear layers. Back-propagation algorithm \cite{rumelhart1985learning}
is usually applied to optimize the parameters of the linear transforms and the non-linearities within this highly non-linear and non-convex
system. With the help of massive labeled training data and powerful
Graphics Processing Units (GPU), DNNs have achieved outstanding performance
in many signal processing and computer vision tasks. However, the working of DNNs is still not completely clear. The
optimized DNNs are usually treated as black-box systems. Some natural
questions are what are the functions of the linear transform and the non-linearities and what is the role of the ``cascade'' in DNNs. 

Some recent works have tried to provide insights into the workings
of DNNs. Bruna and Mallat \cite{mallat2012group,bruna2013invariant}
proposed a Scattering Convolutional Network (SCN) by replacing the
learned filters with wavelet-like transforms. SCN provides feature
representations which are translation and rotation invariant. Zeiler
and Fergus \cite{zeiler2014visualizing} proposed a deconvolution
technique to visualize the intermediate feature layers of a Convolutional
Neural Network (CNN) trained for image classification. The filters
in the first layer are Gabor like, and the deeper layer feature maps
tend to be active only for certain objects. In \cite{lei2018geometric},
the authors suggested that an auto-encoder partitions the low-dimensional
data manifold into cells and approximates each piece by a hyper-plane.
The encoding capability limit of a DNN is described by the upper bound
of the number of cells. Montufar \textit{et al.} \cite{montufar2014number}
have shown that the number of cells is exponentially larger than the
degrees of freedom. Giryes \textit{et al.} \cite{giryes2016deep}
theoretically analyzed the fully connected DNN with i.i.d. random
Gaussian weights. They prove that a DNN with random Gaussian weights
performs a distance-preserving embedding of the data.

{The sparse representation framework \cite{elad2010sparse} 
can generate sparse features for input signals with respect to a given dictionary and sparse representation over an over-complete dictionary can serve as an effective and robust tool in both classification and regression problems.} Depending on the way the signal is modelled, the sparse representation model can be divided into synthesis or analysis model \cite{elad2007analysis}. 

A synthesis model \cite{elad2007analysis} represents a signal $\bm{x} = \bm{D\gamma}$
as a linear combination of a small number of column atoms from a redundant
synthesis dictionary $\bm{D}\in\mathbb{R}^{n\times m}$
with $n<m$ and $\left\Vert \bm{\gamma}\right\Vert _{0} = k \ll m$. 
Sparse pursuit is to seek the sparsest representation $\bm{\gamma}$ given the input
signal $\bm{x}$ and the dictionary $\bm{D}$. Sparse
pursuit algorithms include greedy algorithms \cite{pati1993orthogonal,dai2009subspace,blumensath2009iterative}
and convex relaxation based algorithms \cite{tibshirani1996regression,chen2001atomic,daubechies2004iterative,beck2009fast}. {The greedy algorithms, like Orthogonal Matching Pursuit (OMP)
\cite{pati1993orthogonal}, find at each iteration a sparse coefficient with the aim of reducing the approximation error.} Convex relaxation algorithms
relax the non-convex $l_{0}$ norm to a convex $l_{1}$ norm. The sparse representation problem can then
be solved using basis pursuit (BP) \cite{chen2001atomic} or iterative algorithms, like Iterative Soft-Thresholding
Algorithm (ISTA) \cite{daubechies2004iterative,beck2009fast}.
{The synthesis dictionary learning algorithms \cite{engan1999method,aharon2006k} mainly take an alternating minimization strategy and iterate between a sparse-coding stage and a dictionary update stage. There also exists one-stage synthesis dictionary learning algorithm e.g. \cite{seghouane2018consistent}.}
With well established theories and algorithms, sparse representation over redundant synthesis model has been extensively used in signal and image processing.

{The analysis model \cite{elad2007analysis,rubinstein2013analysis} has also attracted increasing research interests.} A redundant analysis dictionary $\bm{\Omega}\in\mathbb{R}^{m \times n}$ is a tall matrix with $m>n$ where each row of $\bm{\Omega}$ is an atom of the dictionary. The expectation is that the analysis coefficients $\bm{\Omega x}$ are sparse. This means that the analysis
dictionary should be able to sparsify the input signal, whilst preserving its essential information. The analysis dictionary usually serves as a regularization term $\lambda||\bm{\Omega x}||_1$ in the optimization formulation and models the co-sparse prior which can be considered as an extension of the Total Variation (TV) prior. Alternating minimization strategy
can also be applied for learning analysis dictionaries \cite{rubinstein2013analysis,yaghoobi2013constrained,dong2015analysis,ravishankar2013learning,pfister2018learning}. 
Analysis K-SVD \cite{rubinstein2013analysis} iterates between an analysis pursuit operation and a K-SVD dictionary update stage. Yaghoobi \textit{et al.} \cite{yaghoobi2013constrained} proposed a uniformly-normalized tight frame constraint for learning analysis operators.
Analysis Simultaneous Codeword Optimization (ASimCO) algorithm \cite{dong2015analysis} enforces a $K$-sparse constraint on the sparse-coding stage and updates multiple dictionary atoms simultaneously in the dictionary update stage.
Sparsifying transform learning \cite{ravishankar2013learning, pfister2018learning} proposed to constrain the analysis operator to be full rank and well-conditioned.
The GeOmetric Analysis Operator Learning (GOAL) algorithm \cite{hawe2013analysis,kiechle2015bimodal} learns the analysis dictionary by employing an alternative optimization strategy. It
performs dictionary learning on manifolds by minimizing an objective function which
promotes sparse representation and also imposes full rank constraint. 

Building a deep model using sparse representation over redundant synthesis dictionaries has facilitated interpretations of DNNs. 
Rubinstein and Elad \cite{rubinstein2014dictionary} proposed an Analysis-Synthesis Thresholding (AST) model for image deblurring which consists of an analysis dictionary, element-wise hard-thresholding operators and a synthesis dictionary. The AST model can be interpreted
as a fully connected DNN which uses hard-thresholding as non-linearity and has one hidden layer. 
The Double-Sparsity model \cite{rubinstein2010double}
proposes to learn a two-layer synthesis dictionary. The first layer is a dictionary that models a fixed basis, while the second one
is a sparse dictionary with sparse atoms. The effective
dictionary is therefore the multiplication between the two dictionaries. The Double-Sparsity model provides a more efficient and adaptive modeling
and enables learning large dictionary from high-dimensional data {(see also \cite{seghouane2017basis})}.
{A Multi-Layer Convolutional Sparse Coding (ML-CSC) model \cite{papyan2017convolutional,sulam2017multi} is an extension of convolutional sparse coding model \cite{garcia2018convolutional,CAOL_Chun,CAOP_data_Chun} and gives a new
interpretation on the working of Convolutional Neural Networks
(CNNs).} The linear models in CNNs are interpreted as synthesis dictionaries
with convolutional structure and the function of the non-linearities
is interpreted as a simplified sparse pursuit procedure. The ML-CSC model has multiple layers of synthesis dictionaries where
the first dictionary is non-sparse while the following
dictionaries are sparse. {In \cite{CAOL_Chun,CAOP_data_Chun}, the relationship between convolutional analysis operator learning and CNNs is clarified.}
Tariyal \textit{et al.} \cite{tariyal2016deep} proposed a greedy layer-wise deep dictionary
learning method which performs synthesis dictionary learning layer-by-layer.
A parametric approach is proposed in \cite{mahdizadehaghdam2018deep}
to learn a deep dictionary for image classification tasks. The proposed
dictionary learning method contains a forward pass which performs
sparse coding with the given synthesis dictionaries and a backward
pass which updates the dictionaries by gradient descent. 

\begin{figure*}[hbt!]
    \centering
    \includegraphics[width=0.6\linewidth]{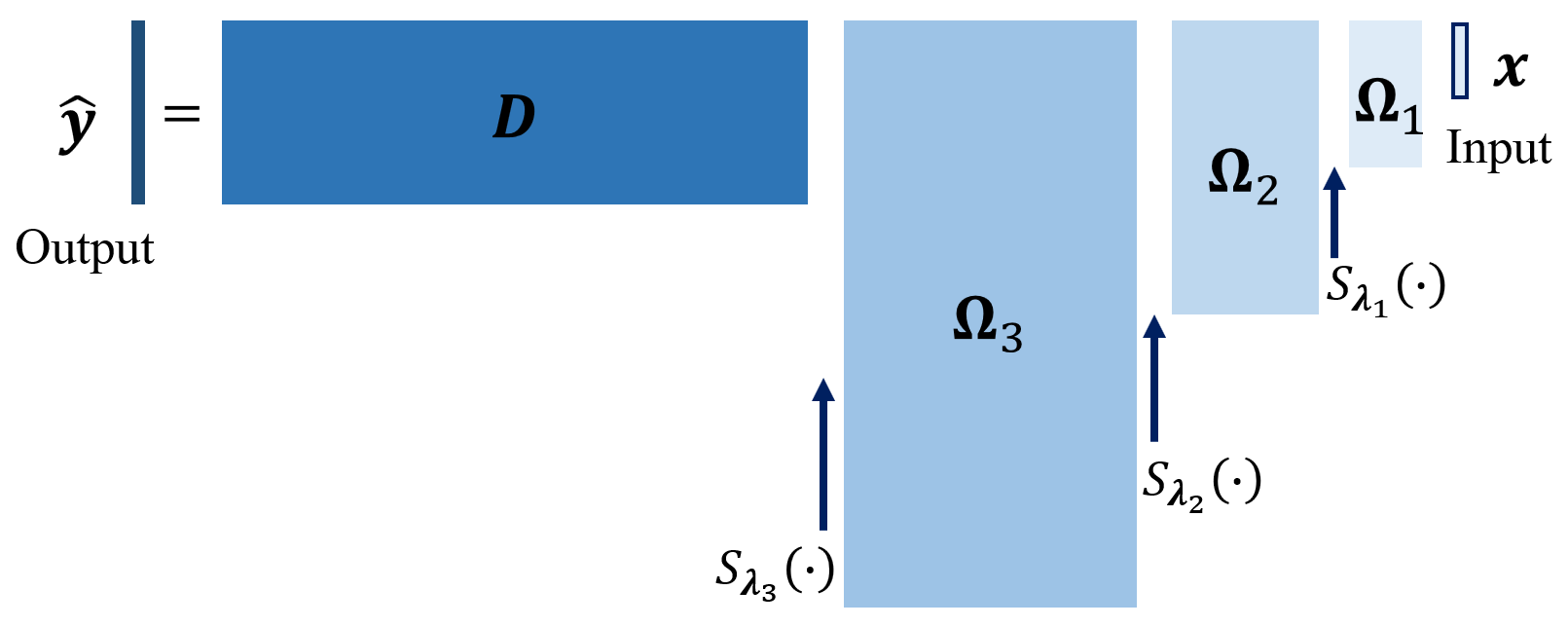}
    \caption{A 3-layer deep analysis dictionary model.
There are 3 layers of analysis dictionaries $\left\{ \bm{\Omega}_{i}\right\} _{i=1}^{3}$ with element-wise soft-thresholding
operators $\left\{ \mathcal{S}_{\bm{\lambda}_{i}}\left(\cdot\right)\right\} _{i=1}^{3}$ and a layer of synthesis dictionary $\bm{D}$. The output signal $\widehat{\bm{y}}$ is obtained through a cascade of matrix multiplications and soft-thresholding operations with input signal $\bm{x}$.}
    \label{fig:overall}
\end{figure*}

The contribution of this paper is two-fold:
\begin{itemize}
  \item We propose a Deep Analysis dictionary Model (DeepAM) which is composed of multiple layers of \textit{analysis} dictionaries with associated soft-thresholding operators and a layer of synthesis dictionary. We propose to characterize each analysis dictionary as a combination of two sub-dictionaries: an Information Preserving Analysis Dictionary (IPAD) and a Clustering Analysis Dictionary (CAD). The IPAD together with the soft-thresholding operator preserves the key information of the input data, and the thresholds are set essentially to denoise the data. The CAD with the associated soft-thresholding operator generates a discriminative representation, and the thresholds are set to facilitate such discrimination.
  \item We propose learning algorithms for DeepAM. To achieve the two different goals of IPAD and CAD, different learning criteria are introduced. {The IPAD learning uses an unsupervised setup, while the CAD learning uses a supervised setup.} We validate our proposed DeepAM on the single image super-resolution task. Simulation results show that the proposed deep dictionary model {is more robust and flexible than a DNN which has the same structure but is optimized using back-propagation. In the self-example image super-resolution task which has limited number of training samples, our proposed DeepAM method achieves significant better performance. Moreover, on noisy image super-resolution, the learned DeepAM can be well adapted to unseen noise levels by rescaling the IPAD and CAD thresholds of the first layer and outperforms DNNs when training and testing noise levels are mismatched.}
\end{itemize}

{Parts of this work were presented in \cite{huang0418_DDSR,huang0519_DDM_Key}.}

The rest of the paper is organized as follows. Section II 
gives an overview of the proposed deep analysis dictionary model. Section III analyzes the proposed model and explains the rationale behind splitting each analysis dictionary into an information preserving and a clustering sub-dictionary.
Section IV introduces the learning algorithm for the deep analysis dictionary model. Section \ref{Results} presents simulation results on single image super-resolution task and Section VI concludes the paper. 

\section{Overview}
We begin this section by briefly introducing the single image super-resolution problem and some notations. We then outline the structure of our deep dictionary model.

    \subsection{Image Super-Resolution}
    The task of single image super-resolution (SISR) is to estimate a (high-resolution) HR image $\widehat{\bm{Y}}$ from an observed (low-resolution) LR image $\bm{X}$. 

    We use patch-based single image super-resolution as the sample application to validate our proposed architecture. 
    Instead of estimating the HR image as a whole, the patch-based approaches \cite{yang2010image,zeyde2010single,timofte2013anchored,timofte2014a+,huang2015fast,huang1215_learning,huang0717_SRHRF+} divide the LR image into overlapping patches and infer a HR patch for each LR patch. The HR image can then be reconstructed using the estimated HR patches by patch overlapping. Learning-based approaches \cite{yang2010image,zeyde2010single,timofte2013anchored,huang2015fast,huang1215_learning,huang0717_SRHRF+,dong2014learning,dong2016image,kim2016accurate} learn the inference model from an external training dataset which contains LR and HR image pairs. The patch-based methods use LR-HR patch pairs $\left\{ \left(\bm{x}_{i}^{0},\bm{y}_{i}\right)\right\} _{i=1}^{N}$
    extracted from the training dataset. The size of the LR patches and the HR patches is assumed to be $p \times p$ and $(s\times p) \times (s\times p)$, respectively. The variable $s$ represents the up-scaling factor. To gain illumination invariance property, the mean value of
    each patch is normally removed. 
    For simplicity, we denote $d_{0}=p^{2}$
    and $d_{L+1}=\left(s\times p\right)^{2}$. By vectorizing the image patches and grouping the training vectors into a matrix, we obtain the input LR training data matrix $\bm{\mathcal{X}}^{0} = [\bm{x}_1^0,\cdots,\bm{x}_N^0] \in\mathbb{R}^{d_{0}\times N}$ and the corresponding ground-truth HR training data matrix $\bm{\mathcal{Y}} = [\bm{y}_1,\cdots,\bm{y}_N] \in\mathbb{R}^{d_{L+1}\times N}$.


\subsection{Deep Analysis Dictionary Model}
To address the SISR problem, we propose a Deep Analysis dictionary Model (DeepAM) which consists of multiple layers of analysis dictionary interleaved with soft-thresholding operations and a single synthesis dictionary. In an $L$-layer DeepAM, there are $L+1$ dictionaries
and $L$ layers that correspond to the non-linear operations. The first $L$ dictionaries are analysis dictionaries and are denoted as $\left\{ \bm{\Omega}_{i} \in \mathbb{R}^{d_{i}\times d_{i-1}}\right\} _{i=1}^{L}$ with $d_{i} \geq d_{i-1}$. The row atoms $\left\{ \bm{\omega}_{i,j}^T \right\}_{j=1}^{d_i}$ of the analysis dictionary $\bm{\Omega}_{i}$ are of unit norm. 
The non-linear operator used in DeepAM is the element-wise soft-threshold\footnote{Soft-thresholding is defined as $\mathcal{S}_{\lambda}(a) = \text{sgn}(a)\max(|a|-\lambda,0).$} $\left\{ \mathcal{S}_{\bm{\lambda}_{i}}\left(\cdot\right)\right\} _{i=1}^{L}$ where $\bm{\lambda}_{i}\in\mathbb{R}^{d_{i}}$ denotes the
threshold vector at layer $i$. 
The  dictionary $\bm{D}$ in the last layer is a synthesis dictionary and is designed to optimize the regression task at hand. 
Fig. \ref{fig:overall} shows an example of a 3-layer deep analysis dictionary model for the image super-resolution task. 

The $L$-layer DeepAM can therefore be expressed mathematically as:
\begin{equation}
    \widehat{\bm{y}} = \bm{D} \mathcal{S}_{\bm{\lambda}_{L}} \left(\bm{\Omega}_{L} \mathcal{S}_{\bm{\lambda}_{L-1}} \left( \cdots \bm{\Omega}_2 \mathcal{S}_{\bm{\lambda}_{1}} \left( \bm{\Omega}_1 \bm{x}^{0} \right) \cdots \right) \right), \label{ddm}
\end{equation}
where $\bm{x}^{0}$ and $\widehat{\bm{y}}$ is the input signal and the estimated output signal, respectively.

Let us denote with $\bm{x}^{i}=\mathcal{S}_{\bm{\lambda}_{i}}(\bm{\Omega}_i\bm{x}^{i-1})$ the output of the $i$-th layer. This means that the input signal $\bm{x}^{i-1}$ is multiplied with the analysis dictionary $\bm{\Omega}_i$ and then passed through the element-wise soft-thresholding operator $\mathcal{S}_{\bm{\lambda}_{i}}(\cdot)$. The thresholded output signal $\bm{x}^{i}$ will be a sparse signal and is expected to be better at predicting the ground-truth signal $\bm{y}$ than $\bm{x}^{i-1}$. 
{Compared to synthesis sparse coding, analysis dictionary with soft-thresholding has much lower computational complexity, since the sparse coding step, which is time consuming, can be omitted during both testing and training}
After $L$ layers, $\bm{x}^{L}$ is transformed to the HR signal domain via the synthesis dictionary. Note that the input LR signal lives in a lower dimensional space when compared to the target HR signal. It is therefore essential for the inference model to be able to non-linearly transform the input data to a higher dimensional space. This is to be achieved by combined use of the analysis dictionaries and the associated soft-thresholding operators.

The proposed DeepAM framework is closely related to Deep Neural Networks (DNNs) with Rectified Linear Unit (ReLU) non-linearity\footnote{{ReLU is defined as $\text{ReLU}(a) = \max(a,0)$.}}. {As ReLU can be considered as the one-sided version of soft-thresholding \cite{papyan2017convolutional, fawzi2015dictionary}, a layer of analysis dictionary with soft-thresholding can be transferred to a layer of Neural Networks with ReLU operator:}
{
\begin{equation}\label{eq:equiv}
    \begin{aligned}
    &\bm{\Omega}_{i+1} \mathcal{S}_{{\bm{\lambda}}_{i}} \left( \bm{\Omega}_i \bm{x}^{i-1} \right) \\
    &= \begin{bmatrix}
    \bm{\Omega}_{i+1} & -\bm{\Omega}_{i+1}
    \end{bmatrix} \text{ReLU} \left (
        \begin{bmatrix}
            \bm{\Omega}_i \\ -\bm{\Omega}_i
        \end{bmatrix}
        \bm{x} - 
        \begin{bmatrix}
            \bm{\lambda}_{i} \\ \bm{\lambda}_{i} 
        \end{bmatrix}
    \right ).
    \end{aligned}
\end{equation}
}

From (\ref{eq:equiv}), we realize that a layer of analysis dictionary and soft-thresholding operation with $n$ atoms can be represented with a layer of neural networks with ReLU and $2n$ neurons. For data which is symmetrically distributed around the origin, DeepAM with soft-thresholding can be more efficient than DNNs with ReLU. This observation will be validated numerically in Section \ref{sec:compareDNNs}.

The learning problem for DeepAM can be formulated as learning the parameters that minimize the mean squared error between the ground-truth data and the estimations:
\begin{equation}
    \underset{\bm{\theta}}{\min}\left\Vert \bm{\mathcal{Y}}-\bm{D}\mathcal{S}_{\bm{\lambda}_{L}}\left(\cdots\bm{\Omega}_{2}\mathcal{S}_{\bm{\lambda}_{1}}\left(\bm{\Omega}_{1}\bm{\mathcal{X}}^{0}\right)\cdots\right)\right\Vert _{F}^{2},\label{eq:train-all}
\end{equation}
where $\bm{\theta} = \left\{ \bm{D},\left\{ \bm{\Omega}_{i}\right\} _{i=1}^{L},\left\{ \bm{\lambda}_{i} \right\} _{i=1}^{L} \right\}$ denotes all the parameters of the $L$-layer DeepAM.

Optimizing (\ref{eq:train-all}) directly can be very difficult as it involves non-convex matrix factorization and the learning of the parameters of the non-linear operators. Standard tools for optimizing DNNs can be utilized, for example back-propagation algorithm \cite{rumelhart1985learning}. However this would lead to effective but difficult to interpret results.

Our goal instead is to build a deep dictionary model with a higher interpretability through imposing specific objectives to different components in the model.
The analysis dictionary and threshold pairs play a key role in DeepAM as they determine the way effective features are generated. The synthesis dictionary instead can be learned using least squares once all the analysis dictionaries and thresholds have been determined. We propose a layer-wise learning strategy to learn the pair of analysis dictionary and soft-thresholding operators. In this way, we can obtain a system where the purpose of every component is easier to interpret.

\section{Analyzing the Deep Analysis Dictionary Model}\label{Analyze}

To justify our approach, we begin by considering a 1-layer DeepAM system:
\begin{equation}\label{eq:singlelayer}
    \widehat{\bm{y}_{i}} = \bm{D} \mathcal{S}_{\bm{\lambda}_1}(\bm{\Omega}_1\bm{x}_{i}^{0}),
\end{equation}
where $\bm{x}_{i}^{0} \in \mathbb{R}^{d_{0}}$ is one of the elements of $\bm{\mathcal{X}}$, $\widehat{\bm{y}_{i}} \in \mathbb{R}^{d_{2}}$ with $d_{2}>d_{0}$, {$\bm{D} \in \mathbb{R}^{d_2 \times d_1}$}, and $\bm{\Omega}_1 \in \mathbb{R}^{d_1 \times d_0}$.

\begin{figure}
    \centering
    \includegraphics[height=0.5\linewidth]{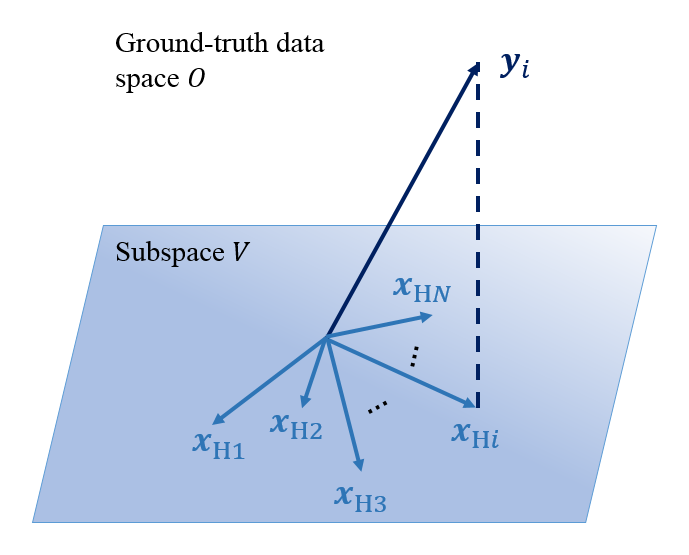}
    \caption{In image super-resolution, the input data spans a low-dimensional subspace $V$ within the HR data space $O$. The objective is to estimate the unknown HR signal $\bm{y}_i$ based on the input LR signal $\bm{x}_{\text{H}i}$. The dashed line represents the residual signal $\bm{r}_i=\bm{y}_i - \bm{x}_{\text{H}i}$ which is orthogonal to the subspace spanned by the input data.}
    \label{fig:subspace}
\end{figure}


A 1-layer DeepAM is similar to the analysis-thresholding synthesis model \cite{rubinstein2014dictionary} and {the single hidden-layer CNN model in iterative neural networks \cite{chun2018deep,chun2019bcd,chun2020momentum}.}
We assume, for the sake of argument, that the degradation model is linear. That is, there exists a degradation matrix $\bm{H} \in \mathbb{R}^{d_{0} \times d_{2}}$ such that $\bm{x}_{i}^{0} = \bm{H} \bm{y}_{i}$.
Denote $\bm{x}_{\text{H}i}=\bm{H}^{\dagger}\bm{x}_{i}^{0} \in \mathbb{R}^{d_2}$ as the projection of the LR signal $\bm{x}^{0}$ onto the HR signal space with the pseudo-inverse matrix $\bm{H}^{\dagger}$. 

As shown in Fig. \ref{fig:subspace}, the signal $\bm{x}_{\text{H}i}$ lies in a low-dimensional subspace $V \subset O$ of the ground-truth HR data space $O$. 
A linear operation will not be able to recover the components that are orthogonal to $V$ (i.e. the dashed line in Fig. \ref{fig:subspace}). 
It is therefore imperative to design the analysis dictionary $\bm{\Omega}_1$ and the non-linear soft-thresholding operation $\mathcal{S}_{\bm{\lambda}_1}(\cdot)$ in a way that facilitates the recovery of the information of $\bm{y}$ in $V^{\bot}$. 

\begin{figure}
    \centering
    \includegraphics[height=0.5\linewidth]{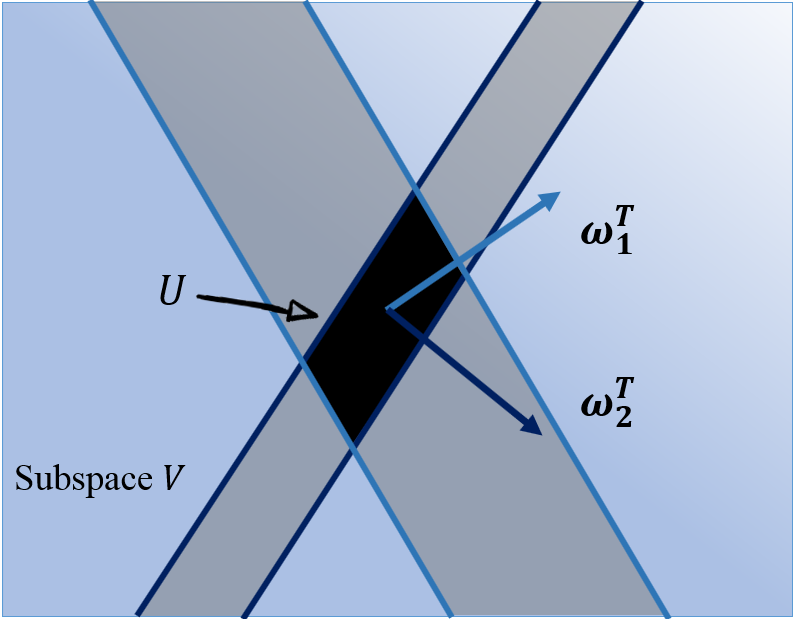}
    \caption{The analysis and soft-thresholding operations partition the input data subspace $V$. There are two pairs of analysis atom and soft-thresholding operator. After soft-thresholding, the data in the gray region is with 1 zero coefficient and the data in the convex polyhedron $U$ (i.e. the black region) is with all zero coefficients.}
    \label{fig:spacePartition}
\end{figure}

When we multiply $\bm{x}^{0}$ with $\bm{\Omega}_1$, the analysis dictionary atoms $\left\{\bm{\omega}_{1,j}^T \right\} _{j=1}^{d_{1}}$ project the input LR data onto specific directions.
After soft-thresholding, the resulting signal $\bm{x}^{1} = \mathcal{S}_{\bm{\lambda}_1}(\bm{\Omega}_1\bm{x}^{0})$ is sparse and the end result is a partitioning of $V$ as shown in Fig. \ref{fig:spacePartition}. 
The input data within each piece is then linearly transformed for prediction, and all the linear transforms are jointly described by the synthesis dictionary $\bm{D}$.

We note that if we assume that all thresholds are large, there is a convex polyhedron $U$ in which all input data will be set to zero by the analysis and the thresholding operations, that is, $\mathcal{S}_{\bm{\lambda}_1}(\bm{\Omega}_1\bm{x}^{0})=\bm{0}$, $\forall \bm{x}^{0} \in U$ (see the central black region in Fig. \ref{fig:spacePartition}). Therefore, the corresponding estimation $\widehat{\bm{y}}$ will be zero, and the information of the data within the convex polyhedron $U$ will then be completely lost. This may lead to a large mean squared error for prediction. 

This suggests that not all thresholds should be too large. The problem can be solved if there is a set of analysis dictionary atoms with small thresholds. If we assume that the signal subspace $V$ has dimension $K$, then in order not to lose essential information there should be at least $K$ pairs of analysis dictionary atoms and soft-thresholds for \textit{information preservation}. These $K$ atoms are associated with $K$ small soft-thresholds and are able to fully represent the input data space. Therefore the $K$ pairs of analysis atoms and soft-thresholds together with the corresponding synthesis atoms provide a baseline estimation for the input data. The remaining analysis dictionary atoms can instead be associated with large thresholds. The outcome of these analysis and thresholding operations is a \textit{clustering} of the input data. 
{That is, the data with the same sparsity pattern can be grouped into a cluster and shares the same set of atoms in the synthesis dictionary for prediction.}
The corresponding synthesis atoms then help recover the signal components within the orthogonal subspace $V^{\bot}$.

\begin{figure}
    \centering
    \includegraphics[height=0.4\linewidth]{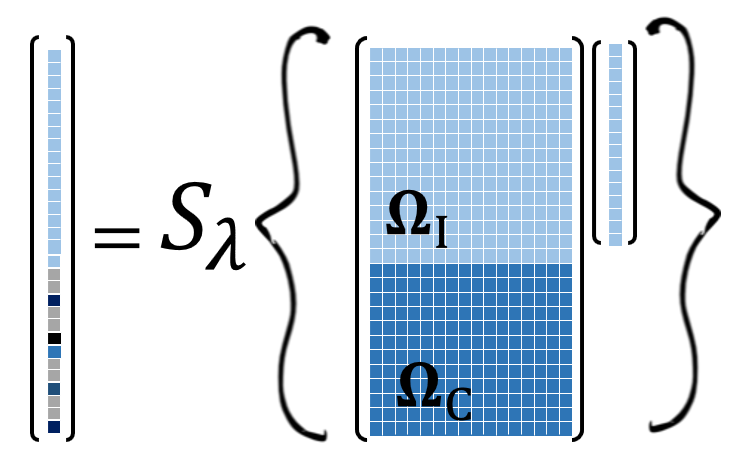}
    \caption{The analysis dictionary $\bm{\Omega}$ is designed to consist of an information preserving analysis dictionary $\bm{\Omega}_{\text{I}}$ and a clustering analysis dictionary $\bm{\Omega}_{\text{C}}$. The soft-thresholds corresponding to $\bm{\Omega}_{\text{C}}$ are much higher than those used for $\bm{\Omega}_{\text{I}}$ and result in a sparser representation.} 
    \label{fig:IPADCAD}f
\end{figure}

Based on the above discussion, we propose to divide an analysis dictionary $\bm{\Omega}$ into two sub-dictionaries $\bm{\Omega} = [\bm{\Omega}_{\text{I}};\bm{\Omega}_{\text{C}}]$, and similarly divide each threshold vector into two parts $\bm{\lambda}=\left[\bm{\lambda}_{\text{I}};\bm{\lambda}_{\text{C}}\right]$. 
The Information Preserving Analysis Dictionary (IPAD) $\bm{\Omega}_{\text{I}}$ with its thresholds $\bm{\lambda}_{\text{I}}$ aims at passing key information of the input signal to the next layer. 
The Clustering Analysis Dictionary (CAD) $\bm{\Omega}_{\text{C}}$ with its threshold $\bm{\lambda}_{\text{C}}$ is to facilitate the separation of key feature in the signal. The CAD and thresholding operators provide a highly non-linear prediction.
Fig. \ref{fig:IPADCAD} shows an analysis dictionary and the soft-thresholding operation with the partition of the IPAD part and the CAD part.

\begin{figure}
    \centering
    \includegraphics[height=0.4\linewidth]{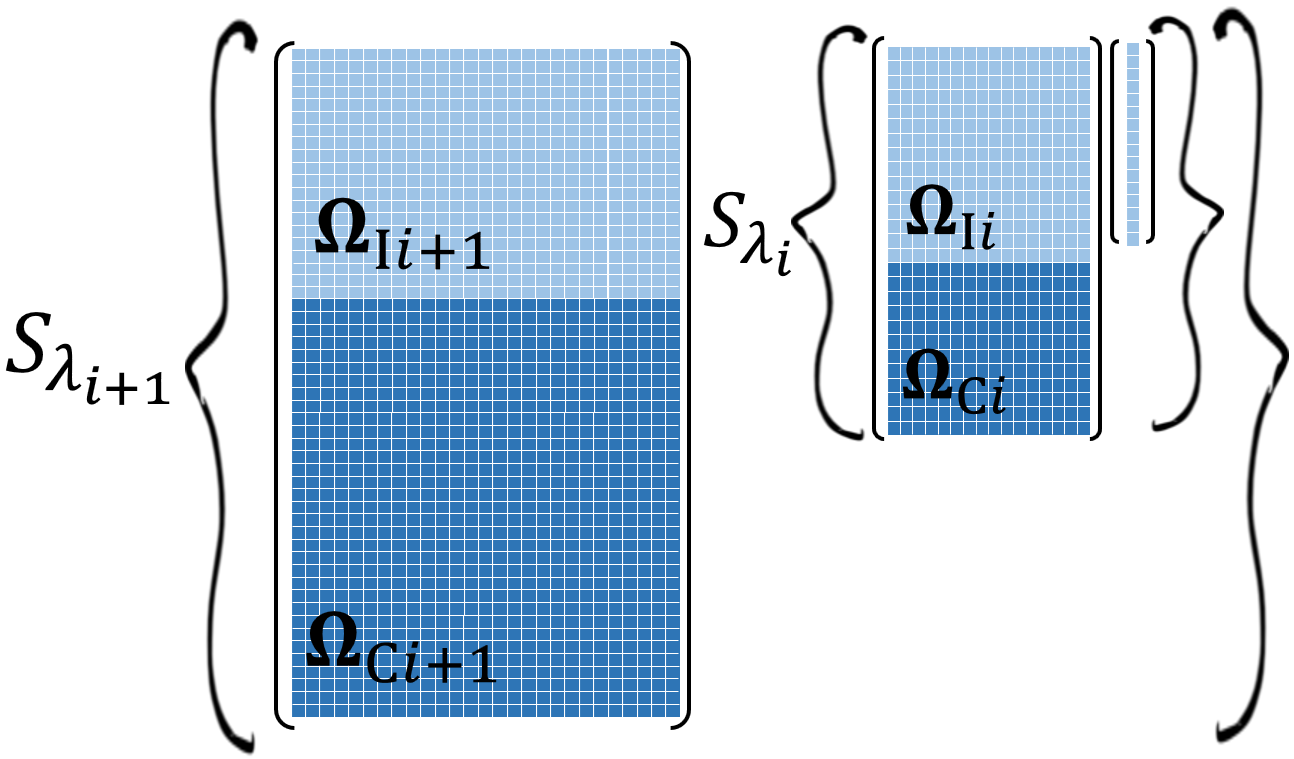}
    \caption{Two consecutive layers in DeepAM. The IPAD and threshold pairs create an information flow channel from input to output, and the CAD and threshold pairs combine information from the previous layer and generate a feature representation that can well represent the residual part.} 
    \label{fig:InfFlow}
\end{figure}

In a multi-layer DeepAM, we adopt the same information preserving and clustering strategy. 
As depicted in Fig. \ref{fig:InfFlow}, the analysis dictionary at each layer is composed of an IPAD part and a CAD part. The IPADs and thresholds $\left\{ ( \bm{\Omega}_{\text{I}i}, \bm{\lambda}_{\text{I}i} ) \right\}_{i=1}^{L}$ create a channel which transmits the information of the input signal to the CAD in each intermediate layer and to the final estimation. The feature representation $\bm{x}_{\text{I}}^L$ generated by the last layer of IPAD and its thresholds should be able to well represent signal components of the HR signal which are within the input data subspace. The CADs and thresholds $\left\{ ( \bm{\Omega}_{\text{C}i}, \bm{\lambda}_{\text{C}i} ) \right\}_{i=1}^{L}$ are the main source of non-linearity in DeepAM. The feature representation $\bm{x}_{\text{C}}^L$ generated by the last layer of CAD should be able to well represent the signal components of $\bm{y}$ which are orthogonal to the input data subspace. 
Therefore, the function of CAD and its thresholds can be interpreted as identifying the data with large energy in the subspace orthogonal to the input data subspace. 
A deep layer of CAD takes the feature representation generated by the IPAD and CAD of the previous layer as input and can generate a non-linear feature representation that cannot be attained by a single layer soft-thresholding with the same number of atoms. Therefore a DeepAM with deeper layers is expected to outperform a shallower one.

\section{Learning a Deep Analysis Dictionary Model}\label{Learn}

In this section, we introduce the proposed learning algorithm for DeepAM. 
In view of the distinctive goals of the two pairs of sub-dictionary and thresholds, different learning criteria have been proposed for learning the IPAD and its thresholds and the CAD and its thresholds.

\subsection{Basic Analysis Dictionary Learning Algorithm}\label{GOAL+}
The IPAD and the CAD have different functions but also share some common learning objectives. 
There are three basic objectives for learning an analysis dictionary: (1) its atoms span a subspace of the input data space; (2) it is able to sparsify the input data; (3) the row atoms are of unit norm. 

Our proposed learning algorithm is an extension of the GeOmetric Analysis Operator Learning (GOAL) algorithm \cite{hawe2013analysis,kiechle2015bimodal} and we denote it as GOAL+. The four learning objectives can be attained by using corresponding constraints. The IPAD and the CAD are learned using modified versions of GOAL+ algorithm.

For simplicity, let us denote the analysis dictionary to be learned as $\bm{\Omega} \in \mathbb{R}^{m \times n}$, the $j$-th atom of $\bm{\Omega}$ as $\bm{\omega}_{j}^T$ and the training data as $\bm{\mathcal{X}} = [\bm{x}_1,\cdots,\bm{x}_N] \in \mathbb{R}^{n \times N}$. We assume that the data $\bm{\mathcal{X}}$ span a $K$ dimensional subspace $V \in \mathbb{R}^{n}$. Let us denote with $\bm{W}\in\mathbb{R}^{n \times K}$ an orthogonal basis of $V$, and with $\bm{U}\in\mathbb{R}^{n \times (n- K)}$ the orthogonal basis of the orthogonal complement $V^{\bot}$. {$\bm{W}$ is obtained by concatenating the leading $K$ left singular vectors of $\bm{\mathcal{X}}$, and $\bm{U}$ is given by the remaining $n-K$ left singular vectors.}

The first learning objective is that the learned analysis dictionary $\bm{\Omega}$ should span only the subspace $V$. 
There are two main reasons for this requirement.
First, the analysis dictionary $\bm{\Omega}$ which spans $V$ can fully preserve the information of the input data.
Second, if an atom $\bm{\omega}^T$ belongs to $V^{\bot}$, it is an unnecessary atom. This is because the coefficients $\bm{\omega}^T\bm{\mathcal{X}}$ will be zero since $V^{\bot}$ is in the null-space of $\bm{\mathcal{X}}$.
We apply a logarithm determinant (log-det) term $h\left(\bm{\Omega}\right)$ to impose the information preservation constraint:
\begin{equation}
    h\left(\bm{\Omega}\right)=-\frac{1}{K\log K}\log\det\left(\frac{1}{m}\bm{W}^{T}\bm{\Omega}^{T}\bm{\Omega}\bm{W}\right).\label{eq:log-det}
\end{equation}

Together with the unit norm constraint, the feasible set of the analysis dictionary $\bm{\Omega}$ is therefore defined as $\Theta=\mathbb{S}_{n-1}^{m}\cap\bm{U}^{\perp}$ with $\mathbb{S}_{n-1}$ being the unit sphere in $\mathbb{R}^{n}$ and $\mathbb{S}_{n-1}^{m}$ being the product of $m$ unit spheres $\mathbb{S}_{n-1}$. In other words the feasible set $\Theta$ restricts the learned atoms in $\bm{\Omega}$ to be of unit norm and excludes the contributions from $\bm{U}$.
(\ref{eq:log-det}) is a generalization of the log-det constraint term applied in GOAL \cite{hawe2013analysis,kiechle2015bimodal}. This is because in our case, $\bm{W}$ defines a basis of the input data space whose size $K$ could be much smaller than the dimension of the input signal $n$ in particular when considering dictionaries in deeper layers. In GOAL \cite{hawe2013analysis,kiechle2015bimodal}, $\bm{W}$ defines a much larger subspace and is with $K=n$ or $K=n-1$.

We achieve the constraint $\bm{\Omega}^{T}\in\Theta$ by performing orthogonal projection onto the tangent space 
$\mathcal{\text{T}}_{\bm{\bm{\Omega}}}(\Theta)$ 
of the manifold $\Theta$ at location $\bm{\Omega}$.
For a row atom $\bm{\omega}^{T}$, the operation of the orthogonal projection onto the tangent space $\mathcal{\text{T}}_{\bm{\bm{\omega}}}(\Theta)$ can be represented by the projection matrix $\bm{P_{\omega}}$ \cite{kiechle2015bimodal}:
\begin{equation}\label{eq:projMat}
    \bm{P_{\omega}}=\mathbf{I}_n-\bm{Q}_{\bm{\omega}}^{\dagger}\bm{Q}_{\bm{\omega}},
\end{equation}
where $\mathbf{I}_n\in\mathbb{R}^{n\times n}$ is the identity
matrix, and $\bm{Q}_{\bm{\omega}}=[2\bm{\omega},\bm{U}]^{T}\in\mathbb{R}^{(n-k+1)\times n}$. 

Sparsifying ability is essential for both IPAD and CAD. 
The sparsifying constraint is imposed by using a log-square function which is a good approximation of the $l_0$ norm:
\begin{equation}\label{eq:sparsify}
    g(\bm{\Omega}) = \frac{1}{Nm \log(1+\nu)}\sum_{i=1}^N \sum_{j=1}^{m}\log \left(1 + \nu (\bm{\omega}_{j}^T \bm{x}_{i})^2 \right),
\end{equation}
where $\nu$ is a tunable parameter which controls the sparsifying ability of the learned dictionary.

\begin{algorithm}[t]
    \SetAlgoLined
    \textbf{Input:} Training data , row number of the dictionary;
    
    \textbf{Initialize:} Initialized $\bm{\Omega}^{(0)}$, $t=0$\;
    \While{halting criterion false}{
        $t \leftarrow t + 1$ \;
         
        Compute gradient of the objective function $\nabla f(\bm{\Omega}^{(t)})$;

        Orthogonal project $\nabla f(\bm{\Omega}^{(t)})$ onto the tangent space of manifold $\Theta$ at $\bm{\Omega}^{(t)}$: $\mathcal{G}\doteq\Pi_{\mathcal{\text{T}}_{\bm{\Omega}^{(t)}}(\Theta)}(\nabla f(\bm{\Omega}^{(t)}))$;

        Find new search direction $\mathcal{H}^{(t)} = -\mathcal{G}+\beta^{(t)}\mathcal{T}_{\mathcal{H}^{(t-1)}}$;
        
        Update $\bm{\Omega}^{(t+1)}$ along the search direction $\mathcal{H}^{(t)}$ using backtracking line search.
    }
    \textbf{Output:} Learned analysis dictionary $\bm{\Omega}$.
 \caption{GOAL+ Algorithm}
\end{algorithm}

{
The combination of the information preservation constraint in (\ref{eq:log-det}) and sparsifying constraint in (\ref{eq:sparsify}) leads to the objective function of GOAL+:}
\begin{equation}
    \bm{\Omega}=\arg\underset{\bm{\Omega}^{T}\in\Theta}{\min}f(\bm{\Omega}),\label{eq:GOAL+}
\end{equation}
{where $f(\bm{\Omega})=g(\bm{\Omega})+\kappa h(\bm{\Omega})$ with $\kappa$ being the regularization parameter.}

The objective function defined in (\ref{eq:GOAL+}) is optimized using a geometric conjugate gradient descent method \cite{absil2009optimization,hawe2013analysis} {which has a good convergence property and with low computational complexity}. The analysis dictionary learning algorithm GOAL+ is summarized in \textbf{Algorithm 1}. 
At iteration $t$, the gradient of the objective function $\nabla f(\bm{\Omega}^{(t)})$ is computed and orthogonal projected on the tangent space of the manifold $\Theta$ at location $\bm{\Omega}^{(t)}$. 
The orthogonal projection of $\nabla f(\bm{\Omega})$ onto the tangent space $\mathcal{\text{T}}_{\bm{\bm{\Omega}}}(\Theta)$ can be expressed as $\Pi_{\mathcal{\text{T}}_{\bm{\bm{\Omega}}}(\Theta)}(\nabla f(\bm{\Omega}))=[\bm{P}_{\bm{\omega}_{1}}\nabla f(\bm{\omega}_{1}),\cdots,\bm{P}_{\bm{\omega}_{d_{\text{I}i}}}\nabla f(\bm{\omega}_{d_{\text{I}i}})]$. Let us denote $\mathcal{G}\doteq\Pi_{\mathcal{\text{T}}_{\bm{\Omega}^{(t)}}(\Theta)}(\nabla f(\bm{\Omega}^{(t)}))$, the search direction can be set as $\mathcal{H}^{(t)}=-\mathcal{G}$. In practice, the search direction is a combination of $\mathcal{G}$ and the previous search direction $\mathcal{T}_{\mathcal{H}^{(t-1)}}$. The updated analysis dictionary $\bm{\Omega}^{(t+1)}$ is then obtained by gradient descent with backtracking line search along the search direction $\mathcal{H}^{(t)}$.
The halting condition is when the analysis dictionary converges or when a pre-defined maximum number of iterations is reached. 
In summary, our optimization approach is similar to that in GOAL \cite{hawe2013analysis} with the exception of the orthogonal projection step as described in (\ref{eq:projMat}) which represents the constraint introduced by the feasible set $\Theta$.
For a more detailed treatment of the geometric conjugate gradient descent we refer to \cite{absil2009optimization,hawe2013analysis}. 
Now that the overall objectives of GOAL+ have been introduced, we can focus on how to tailor the optimization in (\ref{eq:GOAL+}) to achieve the objectives of IPAD and CAD respectively.

\subsection{Learning IPAD and Threshold Pairs}
\label{sec:IPAD}

The function of the IPAD and threshold pair $(\bm{\Omega}_{\text{I}i},\bm{\lambda}_{\text{I}i})$ is to pass key information of the input data $\bm{\mathcal{X}}^{0}$ to deeper layers. The learned IPADs create a channel that enables the information flow from the input signal to the estimated output signal. {The IPAD learning uses the unsupervised learning setup.}

\subsubsection{IPAD Learning}
The training data for learning $\bm{\Omega}_{\text{I}i}$ is the $i$-th layer input training data $\bm{\mathcal{X}}^{i-1}$ (the $(i-1)$-th layer training data is obtained as $\bm{\mathcal{X}}^{i}= \mathcal{S}_{\bm{\lambda}_{i}}(\bm{\Omega}_{i} \bm{\mathcal{X}}^{i-1})$ for $i\geq1$).
Let us denote the rank of the input training data $\bm{\mathcal{X}}^{0}$ at the first layer as $k_0=\text{rk}(\bm{\mathcal{X}}^{0})$ where $\text{rk}(\cdot)$ outputs the rank of a matrix. The IPAD $\bm{\Omega}_{\text{I}i}\in\mathbb{R}^{d_{\text{I}i}\times d_{i-1}}$ is assumed to have $d_{\text{I}i} \geq \text{rk}(\bm{\mathcal{X}}^{0})$ atoms to ensure that the learned IPAD can well represent the input data subspace. 

By setting the training data as $\bm{\mathcal{X}}^{i-1}$, the $i$-th layer analysis dictionary $\bm{\Omega}_{\text{I}i}$ can be learned using GOAL+. The orthonormal basis $\bm{W}\in\mathbb{R}^{d_{i-1} \times k_0}$ is set to be an arbitrary basis of $\bm{\mathcal{X}}^{i-1}$ that corresponds to the signal subspace of $\bm{\mathcal{X}}^{0}$. The orthogonal basis $\bm{U}\in\mathbb{R}^{d_{i-1} \times (d_{i-1}- k_0)}$ is set to span the orthogonal complement of the subspace spanned by $\bm{W}$.

\subsubsection{Learning the Thresholds for IPAD}\label{IPAD_Thresholds}

Given a learned IPAD $\bm{\Omega}_{\text{I}i}$, the analysis coefficients $\bm{\alpha}^{i}=\bm{\Omega}_{\text{I}i}\bm{x}^{i-1}$ contain sufficient information of $\bm{x}^{i-1}$.
When $\bm{\alpha}^{i}$ is a redundant representation or when the input data $\bm{x}^{i-1}$ is noisy, applying a proper thresholding operation to $\bm{\alpha}^{i}$ can further enhance the robustness of the representation. We propose to apply soft-thresholding with small thresholds to the IPAD analysis coefficients as $\bm{z}=\mathcal{S}_{\bm{\lambda}_{\text{I}i}}(\bm{\Omega}_{\text{I}i}\bm{x}^{i-1})$ and interpret the soft-thresholding operation as a form of denoising.

There are related works in the literature about thresholding for redundant representations \cite{elad2006simple,raphan2008optimal,lin2006bayesian}. Elad \cite{elad2006simple} shows that simple shrinkage is effective for redundant representation and interprets the simple shrinkage as the first iteration for solving Basis Pursuit Denoising (BPDN) problems. Raphan and Simoncelli \cite{raphan2008optimal} proposed a denoising scheme for redundant representations based on Stein's Unbiased Risk Estimator. Lin and Lee \cite{lin2006bayesian} proposed a Bayesian framework for finding the optimal $l_{1}$-norm regularization. 

Let us consider a weighed $l_{1}$-norm regularized minimization problem: 
\begin{equation}
    \underset{\bm{z}}{\min} \frac{1}{2}|| \bm{x}-\bm{\Omega}^{T}\bm{z}||_{2}^{2}+
    \sum_{j=1}^{m} \lambda_{\text{I}j}|z_{j}|,\label{eq:P1}
\end{equation}
where $z_{j}$ is the $j$-th coefficient of the sparse vector $\bm{z}$, and $\lambda_{\text{I}j}$ is the corresponding regularization parameter.

Selecting the soft-threshold $\bm{\lambda}_{\text{I}i}$ is equivalent to finding suitable regularization parameters in (\ref{eq:P1}) as the soft-thresholding operation $\mathcal{S}_{\bm{\lambda}}(\bm{\Omega}\bm{x})$ can be interpreted as the first iteration of the Iterative Soft-Thresholding Algorithm (ISTA) \cite{daubechies2004iterative} for solving (\ref{eq:P1}):
\begin{equation}
    \bm{z}^{(1)} = \mathcal{S}_{\bm{\lambda}} \left( \bm{z}^{(0)} +  \bm{\Omega} \left( \bm{x} - \bm{\Omega}^T\bm{z}^{(0)}\right)\right),
    \label{eq:ISTA}
\end{equation}
where the initial sparse code $\bm{z}^{(0)}=\bm{0}$.

Lin and Lee \cite{lin2006bayesian} proposed a method to choose the optimal regularization parameters based on a Bayesian analysis. They assume that the data $\bm{x}$ is with additive i.i.d. zero mean Gaussian noise with variance $\sigma^2$:
\begin{equation}
    P(\bm{x}|\bm{\Omega}^T,\bm{z},\sigma^{2})=\frac{1}{(2\pi\sigma)^{n/2}}\exp\left(-\frac{1}{2\sigma^{2}}\left\Vert \bm{x}-\bm{\Omega}^{T}\bm{z}\right\Vert _{2}^{2}\right)\label{eq:Gauss}
\end{equation}
and assume a Laplacian distribution prior for the sparse code $\bm{z}$ with parameters $\bm{\lambda}$:
\begin{equation}
    P(\bm{z}|\bm{\lambda})=\prod_{j=1}^{m}\frac{\lambda_j}{2}\exp \left(-\lambda_j|z_{j}|\right).\label{eq:Laplace}
\end{equation}

\begin{figure}
    \centering
    \includegraphics[width=0.7\linewidth]{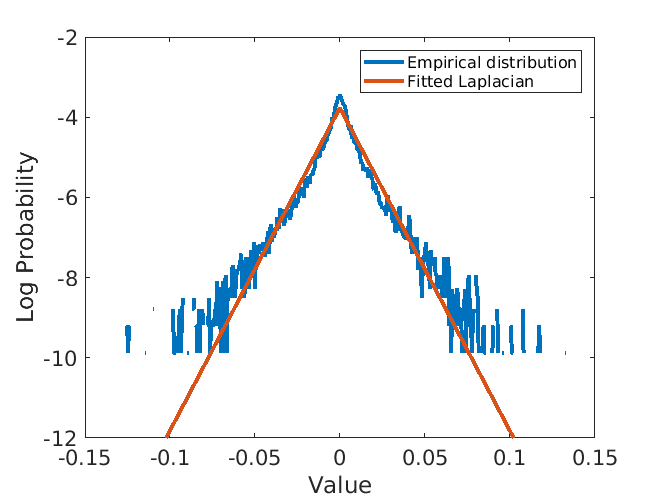}
    \caption{{{An example of the empirical distribution of the response of an atom and the fitted Laplacian distribution.}}}
    \label{fig:laplacian}
\end{figure}

Empirically, we have found that the prior distribution $P\left(\bm{z}\right)$ can be well characterized by an i.i.d. zero-mean Laplacian distribution. 
{Fig. \ref{fig:laplacian} shows an example of the empirical distribution of the inner product between an atom and the input data and the fitted Laplacian distribution.}
Based on the analysis in \cite{lin2006bayesian}, the optimal regularization parameters for (\ref{eq:P1}) can be set as {proportional to the variance of the noise and} inversely proportional to the variance of the Laplacian distribution:
\begin{equation}
    \bm{\lambda} \propto \left[\frac{1}{\sigma_{1}}, \cdots, \frac{1}{\sigma_{m}} \right]^T,\label{eq:para1}
\end{equation}
where $\sigma_{j}$ is the variance of the Laplacian distribution for the $j$-th sparse code $z_{j}$.

From (\ref{eq:para1}), the soft-threshold associated with IPAD $\bm{\Omega}_{\text{I}i}$ is:
\begin{equation}
    \bm{\lambda}_{\text{I}i}=\rho_{\text{I}}\left[\frac{1}{\sigma_{1}},\frac{1}{\sigma_{2}},\cdots,\frac{1}{\sigma_{d_{\text{I}i}}}\right]^{T},
\end{equation}
where $\rho_{\text{I}}$ is a scaling parameter, and the variance $\sigma_{j}$ of the $j$-th coefficient can be estimated using the obtained IPAD $\bm{\Omega}_{\text{I}i}$ and its input data{\footnote{{We estimate the variance $\sigma_j$ of the Laplacian distribution using the mean absolute value of the response of the atom: $\sigma_j = \frac{1}{N} \sum_{k=1}^N \vert \bm{\omega}_{\text{I}i,j}^T \bm{x}^{i-1}_k \vert$.}}}.

There is only a free parameter $\rho$ to be determined. It can
be obtained by solving a 1-dimensional search problem. The optimization problem for $\rho$ is therefore formulated as:
\begin{equation}
    \hat{\rho_{\text{I}}}=\arg\underset{\rho \in \mathcal{D}}{\min}\left\Vert \bm{\mathcal{Y}}-\bm{G}\mathcal{S}_{\rho\bm{\lambda}}\left(\bm{\Omega}_{\text{I}i}\bm{\mathcal{X}}^{i-1}\right)\right\Vert _{F}^{2},\label{eq:1D-opt}
\end{equation}
where $\bm{\lambda}=\left[{1}/{\sigma_{1}},{1}/{\sigma_{2}},\cdots,{1}/{\sigma_{d_{\text{I}i}}}\right]^T$, $\bm{G}=\bm{\mathcal{Y}}\bm{Z}^{T}(\bm{Z}\bm{Z}^{T})^{-1}$ with $\bm{Z}=\mathcal{S}_{\rho\bm{\lambda}}(\bm{\Omega}_{\text{I}i}\bm{\mathcal{X}}^{i-1})$, and $\mathcal{D}$ is a discrete set of values. 

The obtained pair $\left(\boldsymbol{\Omega}_{\text{I}i},\boldsymbol{\lambda}_{\text{I}i}\right)$ should be able to preserve the important information within the input signal and give no worse performance when compared to a linear model without any non-linearity.

\subsection{Learning CAD and Threshold Pairs}
\label{sec:CAD}

The function of the clustering analysis dictionary and threshold pair $(\bm{\Omega}_{\text{C}i},\bm{\lambda}_{\text{C}i})$ is to sparsify its input data and identify the data with large residual energy.
The CAD and threshold pairs at shallow layers provide low-level feature representations for the CADs at deeper layers. 

\subsubsection{CAD Learning}

{The CAD learning uses the supervised learning setup.}
Different from IPAD, learning CAD $\bm{\Omega}_{\text{C}i}$ requires supervision from both the input training data $\bm{\mathcal{X}}^{i-1}$ and the ground-truth HR training data $\bm{\mathcal{Y}}$. 

Let us denote with ${\bm{\mathcal{Y}}}^{i} = \bm{D}_{i}\bm{\mathcal{X}}^{i-1}$ the middle resolution data and with $\bm{\mathcal{E}}^{i}=\bm{\mathcal{Y}} - {\bm{\mathcal{Y}}}^{i}$ the residual data where $\bm{D}_{i} \in \mathbb{R}^{d_{L+1}\times d_{i}}$ is the synthesis dictionary of layer $i$ which minimizes the mean squared reconstruction error and can be obtained by solving:
\begin{equation}
    \bm{D}_{i} = \arg \underset{\bm{D}}{\min} \{ ||\bm{D}\bm{\mathcal{X}}^{i-1}-\bm{\mathcal{Y}}||^2_F + {\Vert\bm{D}\Vert_2^2} \}. 
\end{equation}
It has a closed-form solution given by: 
\begin{equation}
    \bm{D}_i=\bm{\mathcal{Y}}{\bm{\mathcal{X}}^{i-1}}^{T}\left(\bm{\mathcal{X}}^{i-1}{\bm{\mathcal{X}}^{i-1}}^{T} {+\mathbf{I}} \right)^{-1}.\label{eq:synthesis_i}
\end{equation}

The learning objective for CAD $\bm{\Omega}_{\text{C}i}$ is that its atoms should be able to project $\bm{\mathcal{X}}^{i-1}$ onto directions where the data with large residual error has responses with large magnitude.
To achieve that, we propose to first learn an analysis dictionary $\bm{\Psi}_{i} \in \mathbb{R}^{d_{\text{C}i} \times d_{L+1}}$ which acts on the data $\bm{\mathcal{Y}}^{i}$ and is able to jointly sparsify the data ${\bm{\mathcal{Y}}}^{i}$ and the residual data $\bm{\mathcal{E}}^{i}$. That is, the atoms in $\bm{\Psi}_i$ are able to identify the data in $\bm{\mathcal{Y}}^{i}$ with large residual energy.
The $i$-th layer CAD is then re-parameterized as:
\begin{equation}
    \label{eq:repara}
    \bm{\Omega}_{\text{C}i} = \bm{\Psi}_i\bm{D}_{i}.
\end{equation}

As a result, the learned CAD $\bm{\Omega}_{\text{C}i}$ will have the same identification ability as $\bm{\Psi}_i$ since $\bm{\Omega}_{\text{C}i}\bm{x}^{i-1}_{j} = \bm{\Psi}_i\bm{y}^{i}_{j}$. 

We propose the following constraint for learning the analysis dictionary $\bm{\Psi}_i$.
Each analysis atom $\bm{\psi}_{l}^T$ is enforced to be able to jointly sparsify $\bm{\mathcal{Y}}^{i}$ and $\bm{\mathcal{E}}^{i}$:
\begin{equation}\label{eq:sparsify2}
        p(\bm{\Psi}) = c\sum_{j=1}^N \sum_{l=1}^{d_{\text{C}i}}\log \left(1 + \nu \left(\left(\bm{\psi}_{l}^T \bm{y}^{i}_{j})^2 - (\bm{\psi}_{l}^T \bm{e}^{i}_{j}\right)^2\right)^2 \right),
\end{equation}
where $c=1/{Nd_{\text{C}1} \log(1+\nu)}$, $\nu$ is a tunable parameter, and {$\bm{y}^{i}_{j}$ and $\bm{e}^{i}_{j}$ refers to the $j$-th column of $\bm{\mathcal{Y}}^{i}$ and $\bm{\mathcal{E}}^{i}$, respectively}.

The objective function for learning the analysis dictionary $\bm{\Psi}_{i}$ is then formulated as:
\begin{equation}
    \bm{\Psi}_{i}=\arg\underset{\bm{\Psi}^{T}\in\Theta}{\min}f(\bm{\Psi}),\label{eq:GOAL+2}
\end{equation}
{
where $f(\bm{\Psi})=g(\bm{\Psi})+\kappa h(\bm{\Psi}) + \mu p(\bm{\Psi})$ with $\kappa$, and $\mu$ being the regularization parameters. Here, $g(\bm{\cdot})$ and $h(\bm{\cdot})$ are those defined in (\ref{eq:sparsify}) and (\ref{eq:log-det}), respectively.}

The input training data is set to $(\bm{\mathcal{X}}^{i-1},\bm{\mathcal{Y}})$.
Let us denote the rank of $\bm{\mathcal{Y}}$ as  $k_L = \text{rk}(\bm{\mathcal{Y}})$.
The orthogonal basis $\bm{W}\in\mathbb{R}^{d_{L+1} \times k_L}$ is set to be an arbitrary basis of the signal subspace of $\bm{\mathcal{Y}}$. The orthonormal basis $\bm{U}\in\mathbb{R}^{d_{L+1} \times (d_{L+1}- k_L)}$ is set to be a basis spanning the orthogonal complement to the subspace spanned by $\bm{W}$. 

The new objective function in (\ref{eq:GOAL+2}) is optimized using GOAL+ algorithm. With the learned analysis dictionary $\bm{\Psi}_{i}$, the $i$-th layer CAD $\bm{\Omega}_{\text{C}i}$ is obtained as in (\ref{eq:repara}).

\subsubsection{Learning the Thresholds for CAD}

The thresholds for CAD are crucial to the performance of DeepAM as the CAD and threshold pair is the main source of non-linearity in DeepAM. 
The atoms of the learned CAD project the input data onto directions where the data with large residual prediction error will have responses with large magnitude.
After soft-thresholding, the coefficients should be as sparse as possible to achieve a strong discriminative power. 

We propose to set the CAD thresholds as follows:
\begin{equation}
    \bm{\lambda}_{\text{C}i} = \rho_{\text{C}} \left[{\sigma_{1}}, {\sigma_{2}}, \cdots, {\sigma_{d_{\text{C}i}}} \right],\label{eq:para2}
\end{equation}
where $\rho_{\text{C}}$ is a scaling parameter, and $\sigma_{j}$ is the variance of the Laplacian distribution for the $j$-th atom.

As discussed in the previous section, the {analysis sparse codes} can be well modelled by Laplacian distributions. By setting the CAD thresholds proportional to the variance of the analysis coefficients, the proportion of data that has been set to zero for each pair of atom and threshold will be the same. When the synthesis dictionary is applied for reconstruction, the synthesis atoms corresponding to the CAD atoms will be activated with a similar frequency. 

With this simplification, the CAD thresholds can be learned in an efficient way. The scaling parameter $\rho_{\text{C}}$ can be obtained using the same strategy as in (\ref{eq:1D-opt}) by solving a 1-dimensional search problem. 
The optimization problem for $\rho_{\text{C}}$ is formulated as:
\begin{equation}
    {\rho_{\text{C}}}=\arg\underset{\rho \in \mathcal{D}}{\min}\left\Vert {\bm{\mathcal{Y}}}_{R}-\bm{G}\mathcal{S}_{\rho \bm{\lambda}}\left( \bm{\Omega}_{\text{C}i} \bm{\mathcal{X}}^{i-1} \right)\right\Vert _{F}^{2},
    \label{eq:1D-opt_DeepCAMCAD}
\end{equation}
where ${\bm{\mathcal{Y}}}_{R}$ is the estimation residual obtained after using IPAD, $\bm{\lambda}=\left[{\sigma_{1}},{\sigma_{2}},\cdots,{\sigma_{d_{\text{C}i}}}\right]^T$, $\bm{G}=\bm{\mathcal{Y}}_{R}\bm{Z}^{T}(\bm{Z}\bm{Z}^{T})^{-1}$ with $\bm{Z}=\mathcal{S}_{\rho \bm{\lambda}}\left( \bm{\Omega}_{\text{C}i} \bm{\mathcal{X}}^{i-1} \right)$, and $\mathcal{D}$ is a discrete set of values.

In the simulation results in Section \ref{sec:compareDNNs} and \ref{sec:compareSISR}, we will show that the learned CAD thresholds lead to an effective system.

\subsection{Synthesis Dictionary Learning}

The deep analysis dictionary learning can be considered as a layer-wise representation learning process in which the input data $\bm{\mathcal{X}}^{0}$ is consistently non-linearly transformed to a high dimensional feature representation $\bm{\mathcal{X}}^{L}$ with good descriptive and discriminative properties. The synthesis dictionary $\bm{D}$ models the linear transformation from $\bm{\mathcal{X}}^{L}$ to the desired HR counterpart $\bm{\mathcal{Y}}$. Similar to (\ref{eq:synthesis_i}), the synthesis dictionary is learned using least squares:
\begin{equation}
    \bm{D}=\bm{\mathcal{Y}}{\bm{\mathcal{X}}^{L}}^{T}\left(\bm{\mathcal{X}}^{L}{\bm{\mathcal{X}}^{L}}^{T} {+\mathbf{I}}\right)^{-1}.\label{eq:synthesis}
    %
\end{equation}

\subsection{DeepAM Learning Algorithm}

The overall learning algorithm for DeepAM is summarized in \textbf{Algorithm \ref{DeepAMAlg}}. We adopt a layer-wise learning strategy for DeepAM. At each layer, two sub-dictionaries IPAD and CAD are independently learned and then combined to form the analysis dictionary, and the thresholds for IPAD and CAD are obtained with two different strategies. In this way, each pair of analysis dictionary and soft-thresholding operations fulfil two different functionalities. Finally, the synthesis dictionary is learned using least squares. 

\begin{algorithm}[t]\label{DeepAMAlg}
    \SetAlgoLined
    \textbf{Input:} Training data pair $(\bm{\mathcal{X}}^{0},\bm{\mathcal{Y}})$, the number of layers $L$, the structure of DeepAM, {the parameters for learning IPAD $(\nu,\kappa)$, and the parameters for learning CAD $(\nu,\kappa,\mu)$};
    
    \For{$i\gets 1$ \KwTo $L$}{
        Learning $\bm{\Omega}_{\text{I}i}$ using GOAL+ with training data $\bm{\mathcal{X}}^{i-1}$ and objective function (\ref{eq:GOAL+}); 

        Learning $\bm{\Omega}_{\text{C}i}$ using GOAL+ with training data $(\bm{\mathcal{X}}^{i-1},\bm{\mathcal{Y}})$ and objective function (\ref{eq:GOAL+2});  
        
        Learning thresholds $\bm{\lambda}_{\text{I}i}$ and $\bm{\lambda}_{\text{C}i}$;
        
        $\bm{\Omega}_i\gets [\bm{\Omega}_{\text{I}i};\bm{\Omega}_{\text{C}i}]$, $\bm{\lambda}_i\gets [\bm{\lambda}_{\text{I}i};\bm{\lambda}_{\text{C}i}]$;
        
        Update input training data as $\bm{\mathcal{X}}^{i}=\mathcal{S}_{\bm{\lambda}_i}(\bm{\Omega}_i\bm{\mathcal{X}}^{i-1})$;
    }
    Learning the synthesis dictionary $\bm{D}$ as in (\ref{eq:synthesis}).

    \textbf{Output:} Learned DeepAM $\left\{ \left\{ \bm{\Omega}_{i}, \bm{\lambda}_{i}\right\} _{i=1}^{L}, \bm{D} \right\}$.
 \caption{DeepAM Learning Algorithm}
\end{algorithm}

\section{Simulation Results}\label{Results}

In this section, we report the implementation details and numerical results of our proposed DeepAM method and compare our proposed method with Deep Neural Networks learned using back-propagation.

\subsection{Implementation Details}
We use the standard 91 training images \cite{yang2010image} as training dataset and use \textit{Set5} \cite{yang2010image} and \textit{Set14} \cite{zeyde2010single} as the testing datasets. The Peak Signal-to-Noise Ratio (PSNR)\footnote{PSNR=$10\log(\frac{{255}^2}{\sqrt{MSE}})$, where $MSE$ is the mean squared error between the ground-truth HR image and the estimated HR image} is used as the evaluation metric. The color images have been converted from RGB color space to YCbCr color space and image super-resolution is only applied to the luminance component. The low-resolution (LR) images are obtained by down-sampling the ground-truth high-resolution (HR) images by a factor $s=2$ using Matlab function imresize. The size of the low-resolution patches is set to $p=6$ for the purpose of better visualization and easier interpretation. The size of the high-resolution patches is then $12 \times 12$. 

During testing, full patch overlapping is applied to reconstruct the HR images. 
{Instead of predicting a $12 \times 12$ HR patch for each input LR $6 \times 6$ patch, the output of DeepAM is the central $8 \times 8$ HR patch since the input LR patch does not contain sufficient information to predict the boundary pixels\footnote{This is also applied to deep neural network with soft-thresholding operator learned using back-propagation, but not applied to deep neural network with ReLU operator learned using back-propagation due to a degraded performance.}. }

{
\subsection{Parameter Setting}
}
\label{sec:paramsetting}

\begin{figure}
    \centering
    \includegraphics[height=0.24\textwidth]{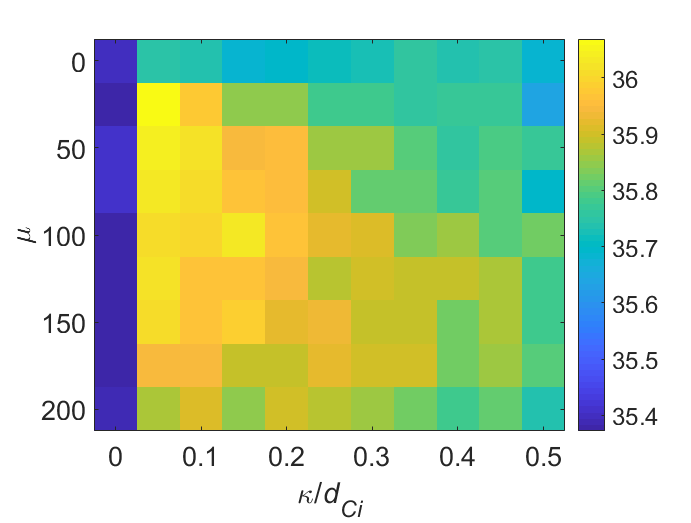}
    \caption{{The performance of 1-layer DeepAM with different combinations of parameters evaluated on \textit{Set5}. All the models are with 1 layer and have 256 atoms.}
     }
    \label{fig:paramSettings}
\end{figure}

\begin{table}
    \center

    \begin{tabular}{|c| C{1.3cm}|C{1.3cm}|C{1.3cm}|}
    \hline 
    {Parameters} & {$\nu$} & {$\kappa$} & {$\mu$}\tabularnewline
    \hline \hline
    {IPAD} & {$100\times d_{i-1}$} & {$d_{\text{I}i}$} & {---}\tabularnewline
    \hline
    {CAD} & {$100\times d_{i-1}$} & {$0.1\times d_{\text{C}i}$} & {100}\tabularnewline
    \hline 
    \end{tabular}{\par}

    \caption{Parameters setting of GOAL+ algorithm for learning the $i$-th layer IPAD $\bm{\Omega}_{\text{I}i} \in \mathbb{R}^{d_{\text{I}i} \times d_{i-1}}$ and CAD $\bm{\Omega}_{\text{C}i} \in \mathbb{R}^{d_{\text{C}i} \times d_{i-1}}$. }
\label{paramSet}
\end{table}

\begin{figure}[t]
    \centering
    \includegraphics[height=0.24\textwidth]{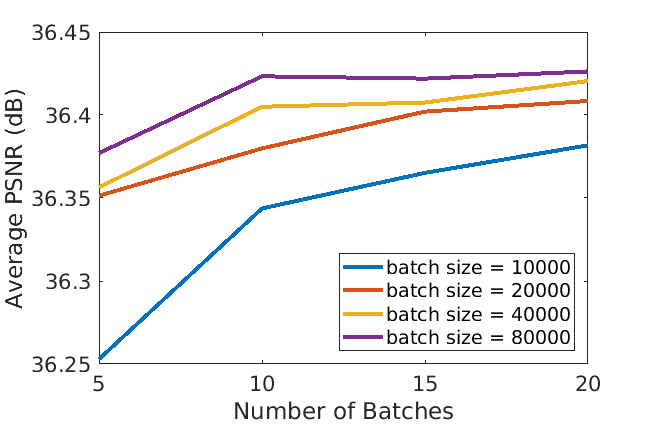}
    \caption{{The performance of learned DeepAM using different batch sizes and number of batches for training evaluated on \textit{Set5}. {All the models are with 3 layer and each layer is with 256 atoms.}}}
    \label{fig:batch}
\end{figure}

\begin{figure}
    \centering
    \includegraphics[width=0.38\textwidth]{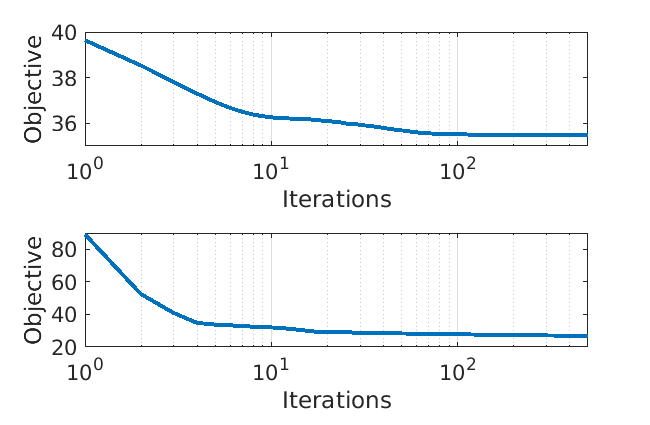}
    \caption{{The objective minimization in IPAD learning of 35 atoms (top) and in CAD learning of 221 atoms (bottom) with the default parameter setting of GOAL+ algorithm.}}
    \label{fig:convergence2}
\end{figure}

\begin{figure*}[hbt!]
	\centering
	\hspace*{\fill}
	\subfigure[$\bm{\Omega}_1 \in \mathbb{R}^{100 \times 36}$.]{
		\label{fig:1layer_1} 
		\includegraphics[height=0.25\textwidth]{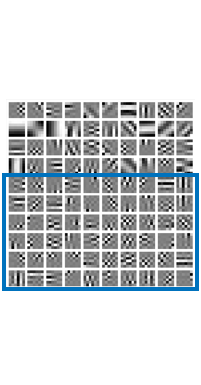}}
		\hfill
	\subfigure[$\bm{\lambda}_1 \in \mathbb{R}^{100}$ and the percentage of data preserved after thresholding.]{
		\label{fig:1layer_2} 
		\includegraphics[height=0.25\textwidth]{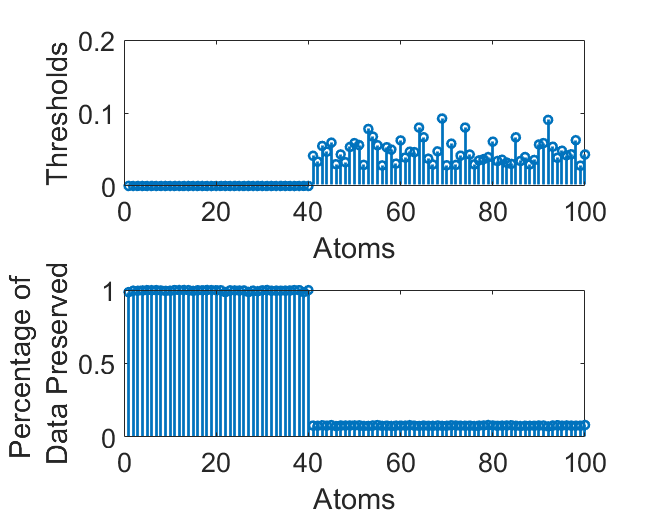}}
		\hfill
	\subfigure[$\bm{D} \in \mathbb{R}^{144 \times 100}$.]{
		\label{fig:1layer_3} 
		\includegraphics[height=0.25\textwidth]{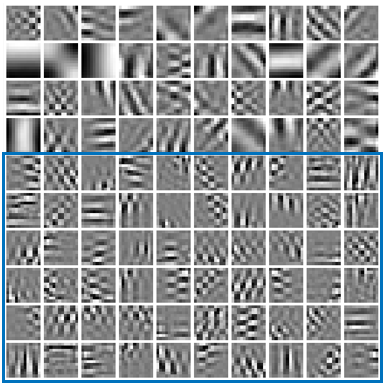}}
	\hspace*{\fill}
	\caption{An example of a learned 1-layer DeepAM. Each atom is displayed
as a 2D patch. The atoms within the blue box are the clustering atoms. In $\bm{\Omega}_1$, the first 40 atoms are the information preserving atoms and the remaining 60 atoms are the clustering atoms.}
    \label{fig:1layer}
\end{figure*}

\begin{figure*}[hbt!]
	\centering
	\hspace*{\fill}
	\subfigure[$\bm{\Omega}_1 \in \mathbb{R}^{100 \times 36}$.]{
		\label{fig:1layer_1bp} 
		\includegraphics[height=0.25\textwidth]{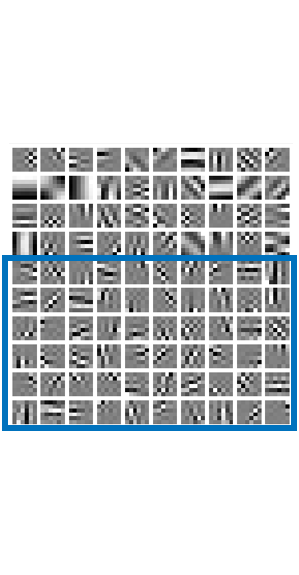}}
		\hfill
	\subfigure[$\bm{\lambda}_1 \in \mathbb{R}^{100}$ and the percentage of data preserved after thresholding.]{
		\label{fig:1layer_2bp} 
		\includegraphics[height=0.25\textwidth]{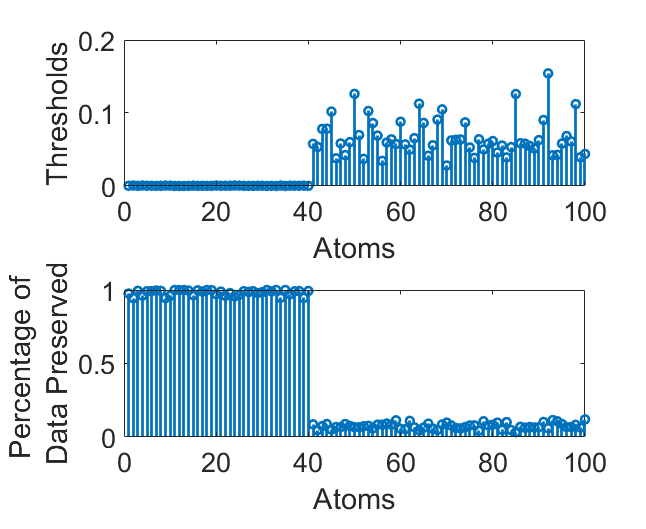}}
		\hfill
	\subfigure[$\bm{D} \in \mathbb{R}^{144 \times 100}$.]{
		\label{fig:1layer_3bp} 
		\includegraphics[height=0.25\textwidth]{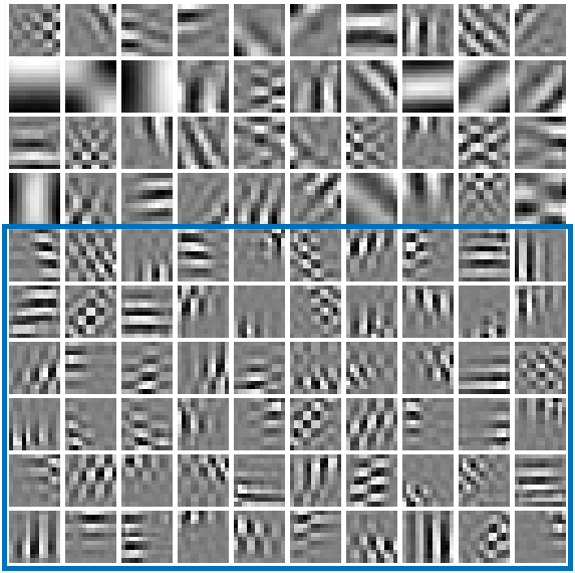}}
	\hspace*{\fill}
	
	\caption{The 1-layer DeepAM further fine-tuned using back-propagation. The dictionary atoms seem more localized. The thresholds are in general larger than those in Fig. \ref{fig:1layer_2}.}
    \label{fig:1layerbp}
\end{figure*}

{In this section, we discuss how several parameters are determined. For IPAD learning, there are two parameters that need to be set: $\nu$ and $\kappa$, and for CAD learning, there are three parameters $\nu$, $\kappa$, and $\mu$. The sparsifying parameter $\nu$ for learning the $i$-th layer IPAD and CAD is set to $100 \times d_{i-1}$ which can lead to learned atoms with good sparsifying ability.}

{As introduced in Section \ref{Learn}, the level of the thresholds for soft-thresholding is determined through a grid search. The discrete set $\mathcal{D}$ defines the searching grid. In this paper, the grid $\mathcal{D}$ is set as $[\cdots,1\times10^{-2},2\times10^{-2},3\times10^{-2},\cdots,1\times10^{-1},2\times10^{-1},3\times10^{-1},\cdots]$. This setting empirically works well. }

{Since the IPAD and its thresholds are responsible for passing essential information from the input, the learned IPAD should span the signal subspace of its input training data. Therefore, a strong rank constraint should be imposed. The rank constraint parameter for IPAD learning is set to $d_{\text{I}i}$.}

{The CAD and its thresholds are responsible for identifying the missing high-frequency information of the input signal. 
Different parameter combinations $(\kappa,\mu)$ have been evaluated to find the best parameter setting for learning CAD. Around 20,000 training LR and HR patch pairs have been extracted from the training dataset. For each combination of $\kappa$ and $\mu$, a 1-layer DeepAM which has 256 atoms in the analysis dictionary is learned using the extracted training data and evaluated on \textit{Set5} \cite{yang2010image}. The number of conjugate gradient descent iterations used for learning IPAD and CAD is set to 500. The halting condition is when the maximum number of iterations is reached.
The evaluation results for different parameter combinations are shown in Fig. \ref{fig:paramSettings}.}

{From Fig. \ref{fig:paramSettings}, we can see that both the rank constraint $h(\cdot)$ and the joint sparsifying constraint $p(\cdot)$ are essential to learn a DeepAM with good performance. When the rank constraint is not used (i.e. the rank constraint parameter $\kappa=0$), the performance of the learned DeepAM is around 35.4 dB and is around 0.6 dB worse than the result of the best parameter combination. 
When the joint sparsifying constraint is not used in the optimization (i.e. the joint sparsifying constraint parameter $\mu=0$), the performance of the learned DeepAM is around 0.3 dB worse than the result of the best parameter combination. 
The rank constraint $h(\cdot)$ and the joint sparsifying constraint $p(\cdot)$ are therefore both important for learning a good CAD. In the rest of the paper, the parameter combination $(\kappa,\mu)$ for learning CAD is set to $(0.1\times d_{\text{C}i}, 100)$.}

{Table \ref{paramSet} summarizes the parameters setting of GOAL+ algorithm for learning the $i$-th layer Information Preserving Analysis Dictionary (IPAD) and Clustering Analysis Dictionary (CAD). Both the IPAD and the CAD are initialized with i.i.d. Gaussian random entries. Fig. \ref{fig:convergence2} shows the objective minimization of the IPAD learning and the CAD learning with the default settings and when learning from around 20,000 training samples.
}

{We further explore the dependency of the performance of learned DeepAM on the training data size.
Fig. \ref{fig:batch} shows the performance of DeepAM when using different batch size and number of batches for training. The total number of training samples is around 80,000. For each batch, 100 iterations of conjugate gradient descent are performed to minimize the objective function. We can find that using large batch size for training is beneficial and learning using around 15 batches is sufficient to achieve a good performance. 
Therefore, in the rest of the paper, we set the batch size as $\min(N,4\times10^4)$ and set the number of batches as 15.}

{Given the default settings determined in this section, the computation time for learning a layer of DeepAM with 256 atoms using 320,000 training data pairs is about 20 minutes on a Intel Core i7-8700 6-core 3.20GHz with 32G RAM CPU machine.}

\subsection{Visualization of the Learned DeepAM}
\label{sec:Visualize}

In this section, we will show and analyze the learned DeepAM with the implementation details as described in the previous section.

\begin{figure}
    \centering
    \includegraphics[width=0.35\textwidth]{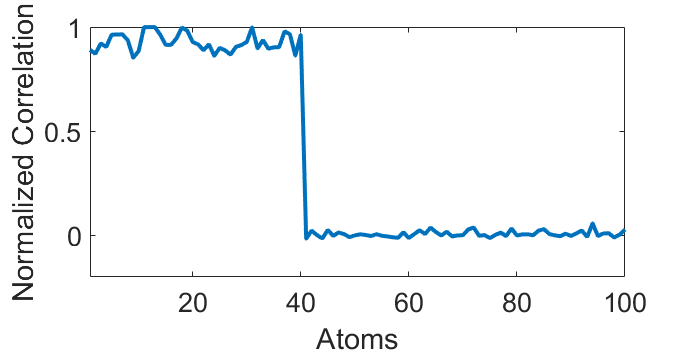}
    \caption{{The normalized inner product between the HR projection of an analysis atom $\bm{H}^{\dagger}\bm{\omega}$ and its corresponding synthesis atom $\bm{d}$.}}
    \label{fig:normalizedCorrel}
\end{figure}

\begin{figure*}[hbt!]
	\centering
	\hspace*{\fill}
	\subfigure[$\bm{\Omega}_1 \in \mathbb{R}^{64 \times 36}$.]{
		\label{fig:2layer_1} 
		\includegraphics[height=0.25\textwidth]{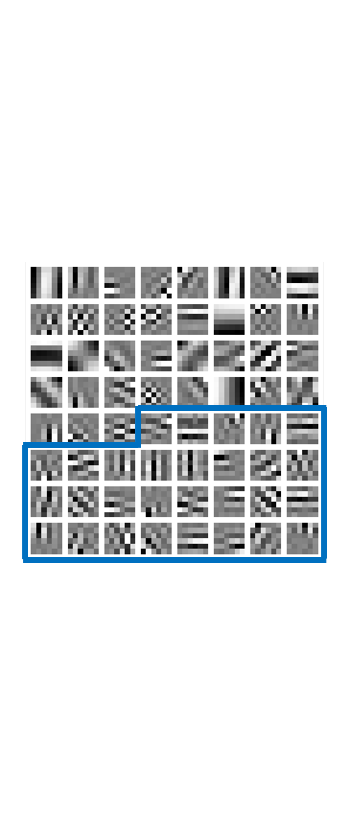}}
		\hfill
	\subfigure[$\bm{\Omega}_2 \in \mathbb{R}^{144 \times 64}$.]{
		\label{fig:2layer_2} 
		\includegraphics[height=0.25\textwidth]{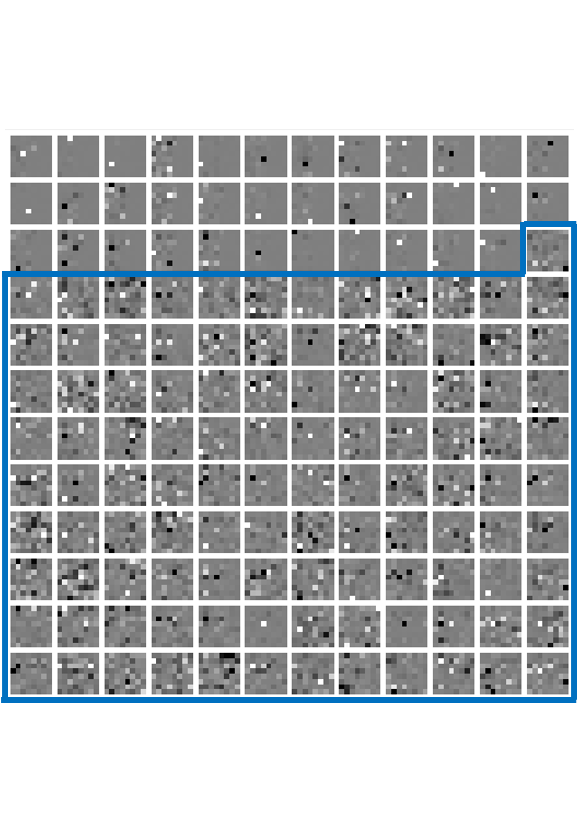}}
		\hfill
	\subfigure[$\bm{\Omega}_2\bm{\Omega}_1 \in \mathbb{R}^{144 \times 36}$.]{
		\label{fig:2layer_3} 
		\includegraphics[height=0.25\textwidth]{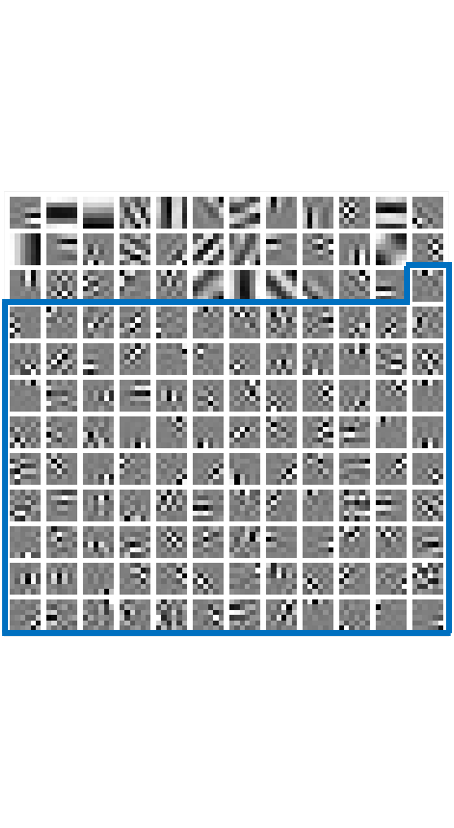}}
		\hfill
	\subfigure[$\bm{D} \in \mathbb{R}^{144 \times 144}$.]{
		\label{fig:2layer_4} 
		\includegraphics[height=0.25\textwidth]{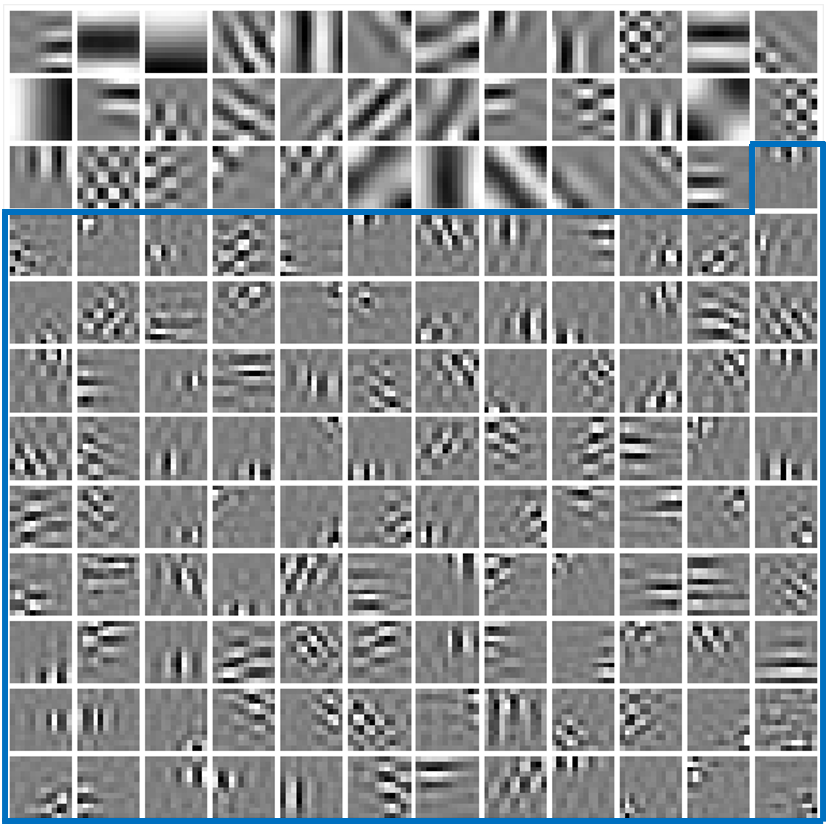}}
	\hspace*{\fill}
	
	\caption{An example of a learned 2-layer DeepAM. Each atom is displayed
as a 2D patch. The atoms within the blue box are the clustering atoms. }
    \label{fig:2layer}
\end{figure*}

\begin{figure*}[hbt!]
	\centering
	\hspace*{\fill}
	\subfigure[$\bm{\Omega}_1 \in \mathbb{R}^{64 \times 36}$.]{
		\label{fig:2layer_1bp} 
		\includegraphics[height=0.25\textwidth]{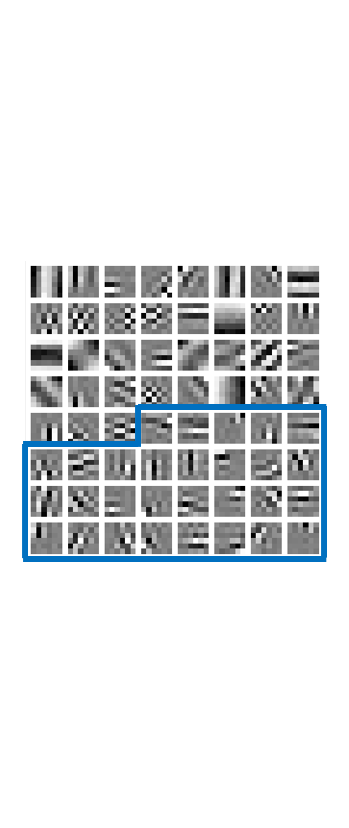}}
		\hfill
	\subfigure[$\bm{\Omega}_2 \in \mathbb{R}^{144 \times 64}$.]{
		\label{fig:2layer_2bp} 
		\includegraphics[height=0.25\textwidth]{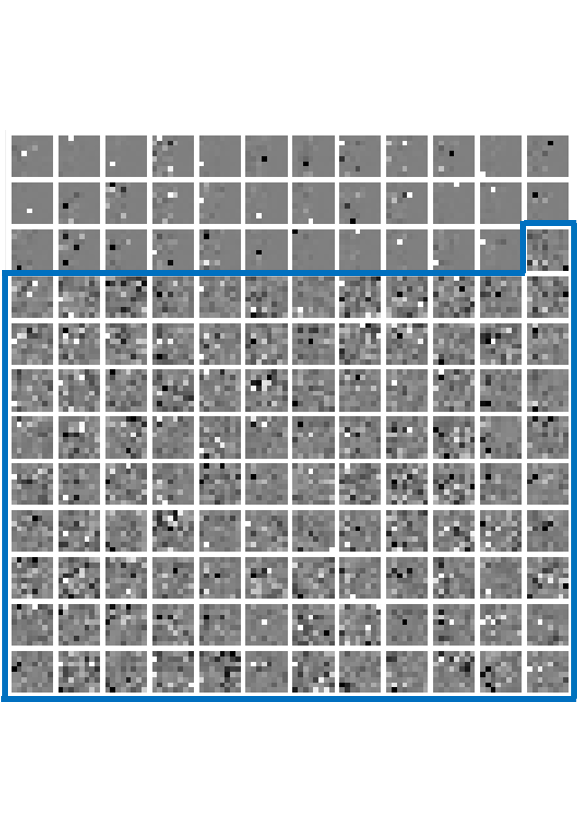}}
		\hfill
	\subfigure[$\bm{\Omega}_2\bm{\Omega}_1 \in \mathbb{R}^{144 \times 36}$.]{
		\label{fig:2layer_3bp} 
		\includegraphics[height=0.25\textwidth]{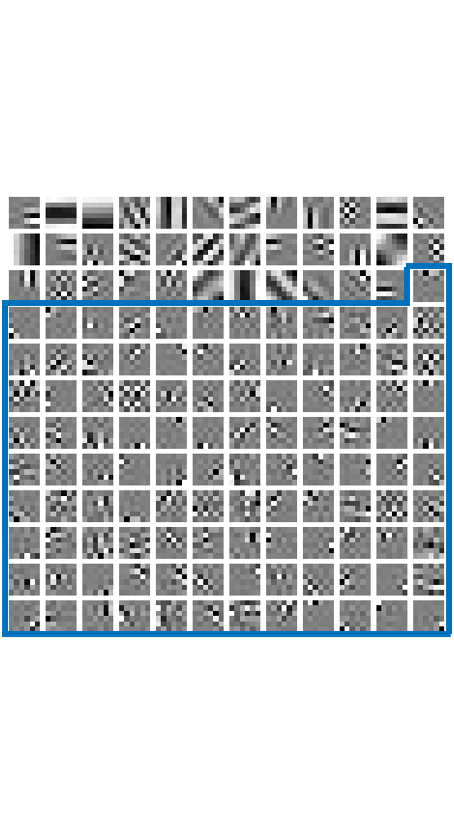}}
		\hfill
	\subfigure[$\bm{D} \in \mathbb{R}^{144 \times 144}$.]{
		\label{fig:2layer_4bp} 
		\includegraphics[height=0.25\textwidth]{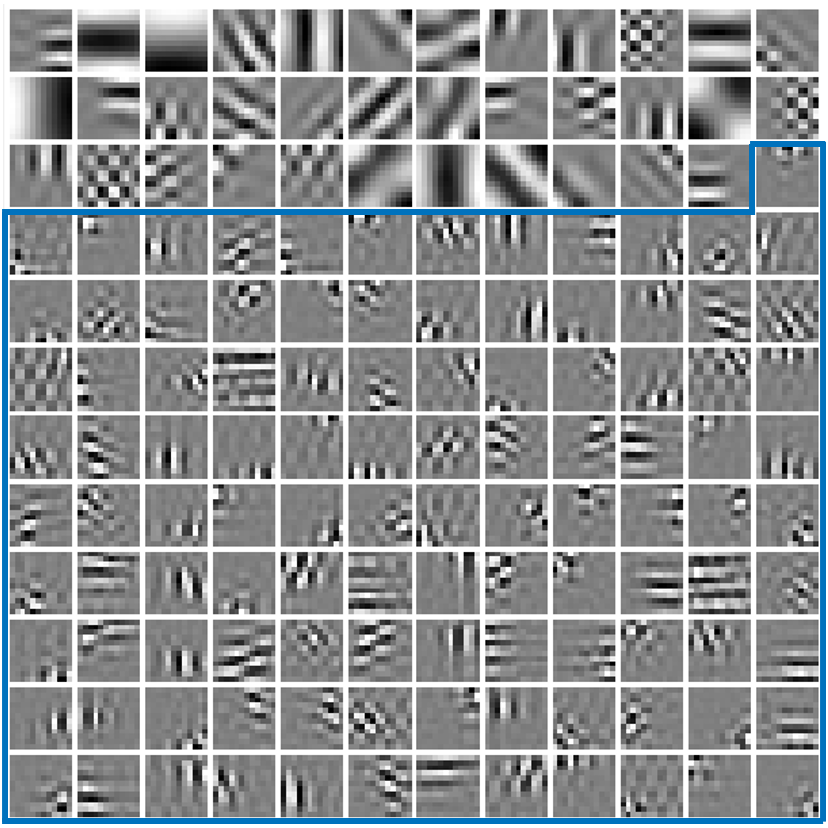}}
	\hspace*{\fill}
	
	\caption{The dictionaries of the 2-layer DeepAM further fine-tuned using back-propagation. }
    \label{fig:2layerbp}
\end{figure*}

Fig. \ref{fig:1layer} shows an example of a learned 1-layer DeepAM. It contains an analysis dictionary $\bm{\Omega}_1$, thresholds $\bm{\lambda}_1$ and a synthesis dictionary $\bm{D}$. Each atom is displayed in a 2D patch in which black and white corresponds to the smallest and the largest value, respectively. The number of the information preserving atoms is set to 40 which is larger than the rank of the input data. The thresholds depicted in Fig. \ref{fig:1layer_2} show a clear bimodal behaviour. The first 40 thresholds are close to zero, while the remaining 60 thresholds are relatively large. After thresholding, almost all coefficients corresponding to IPAD are non-zero, and the percentage of non-zero coefficients of different CAD atoms are similar and are around 8\%. This indicates that modelling the distribution of the analysis coefficients as a Laplacian distribution is a good approximation. The atoms within the blue box are the clustering atoms. The atoms in IPAD shown in Fig. \ref{fig:1layer_1} are similar to the LR versions of their corresponding synthesis atoms in Fig. \ref{fig:1layer_3}. 
The CAD atoms look like directional filters and are more localized. There is little low-frequency information. The corresponding synthesis atoms are correlated to the CAD atoms, however, they are not the HR counterpart. 
{Fig. \ref{fig:normalizedCorrel} shows the normalized inner product between the HR projection of an analysis atom $\bm{H}^{\dagger}\bm{\omega}$ and its corresponding synthesis atom $\bm{d}$. We can see that the analysis atoms and synthesis atoms corresponding to the IPAD part have high inner products, while the inner products between the analysis atoms and synthesis atoms corresponding to the CAD are very small.} This shows that, in line with our objective, the synthesis atoms which correspond to the CAD part are nearly orthogonal to the LR data subspace.

{Back-propagation can be used to further update all the parameters (both dictionaries and thresholds) in our learned DeepAM.} The back-propagation update is implemented using Pytorch with Adam optimizer \cite{kingma2014adam}, batch size $1024$, initial learning rate $10^{-3}$, learning rate decay step $20$, and decay rate $0.1$. The parameter setting has been tuned to achieve the best performance. 

Fig. \ref{fig:1layerbp} shows the 1-layer DeepAM after updating using back-propagation. With back-propagation, the performance of DeepAM has a rapid improvement with the first 5 epochs and converges within 20 epochs. After back-propagation update, the average PSNR evaluated on \textit{Set5} has improved by approximately 0.3 dB. We can find that the different characteristics of the IPAD part and the CAD part are preserved on the updated DeepAM. There are subtle differences on the updated dictionaries. In general, the IPAD atoms have no visible changes, while the CAD atoms have become more localized. The thresholds continue to have a bimodal behaviour. There is only a slight change on the percentage of non-zero coefficients of different atoms. 

\begin{figure}
    \centering
    \includegraphics[width=0.4\textwidth]{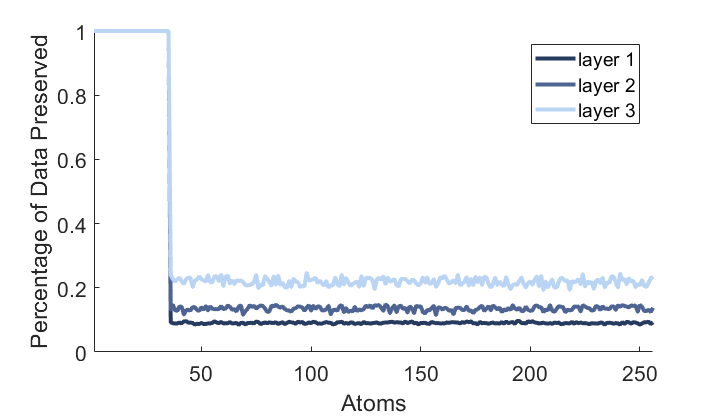}
    \caption{The percentage of data preserved after thresholding for the atoms in 3 different layers of the 3-layer DeepAM in Table \ref{tab:largeTrainSet}.}
    \label{fig:dataPreserved}
\end{figure}

\begin{table*}[t]
    \center
    \begin{tabular}{|c||c|c|c||c|c|c|}
    \hline 
    {Training Data Size} & \multicolumn{3}{| c ||}{5,000} & \multicolumn{3}{| c |}{320,000}\tabularnewline
    \hline 
    {Method} & {DNN-R} & {DNN-S} & {DeepAM} & {DNN-R} & {DNN-S} & {DeepAM} \tabularnewline
    \hline \hline
    {\textit{Set5}} & {35.06} & {35.62} & {\textbf{35.96}} & {36.32} & \textbf{36.62} & {{36.44}}\tabularnewline
    \hline 
    {\textit{Set14}} & {31.36} & {31.63} & {\textbf{31.87}} & {32.01} & \textbf{32.30} & {{32.17}} \tabularnewline
    \hline 
    \end{tabular}{ \par}
    \caption{Average PSNR (dB) by different methods {learned using a small and a large training dataset} evaluated on \textit{Set5} \cite{yang2010image} and \textit{Set14} \cite{zeyde2010single}. All the models are with 3 layers and each layer has 256 atoms. (The best result in each row is in bold.)}
    \label{tab:CompareDNN}
\end{table*}

Fig. \ref{fig:2layer} shows the dictionaries of a learned 2-layer DeepAM including two analysis dictionaries $\bm{\Omega}_1$, $\bm{\Omega}_2$ and a synthesis dictionary $\bm{D}$. The first analysis dictionary $\bm{\Omega}_1$ is similar to that in Fig. \ref{fig:1layer_1}, while its CAD part mainly contains directional filters due to a smaller number of clustering atoms. The second analysis dictionary $\bm{\Omega}_2$ is shown in Fig. \ref{fig:2layer_2} and is a sparse dictionary 
where the sparse atoms can be considered as indicating a weighted combination of the first layer analysis dictionary atoms if the soft-thresholding operation is neglected. The effective dictionary $\bm{\Omega}_{21} = \bm{\Omega}_2\bm{\Omega}_1$ shown in Fig. \ref{fig:2layer_3} can partially show the effective atoms applied to the input LR data whose IPAD part is similar to that in $\bm{\Omega}_1$ and CAD part contains more localized atoms when compared to those in $\bm{\Omega}_1$. 
Similar observations can be found in a deeper analysis dictionary in DeepAM. The synthesis dictionary has similar characteristics as the one in the 1-layer DeepAM. 

Fig. \ref{fig:2layerbp} shows the dictionaries of the 2-layer DeepAM after updating with back-propagation. The back-propagation slightly updates the dictionaries and converges within 20 epochs. The average PSNR evaluated on \textit{Set5} improves by 0.2 dB after the first 5 epochs and achieved a 0.3 dB improvement after convergence. As in the 1-layer DeepAM case, after back-propagation, there is still a clear difference between the IPAD atoms and the CAD atoms. The IPAD atoms did not change significantly, while the CAD atoms in $\bm{\Omega}_1$ and the effective dictionary $\bm{\Omega}_{21}$ have become more localized.

Fig. \ref{fig:dataPreserved} further shows the percentage of non-zero coefficients for each atom in 3 different layers of the 3-layer DeepAM. 
We can find that the percentage of non-zero coefficients has a bimodal behaviour in all three layers which is the same to that shown in Fig. \ref{fig:1layer_2}. After thresholding, the percentage of non-zero coefficients corresponding to CAD atoms are almost the same in each layer. 
The percentage of non-zero coefficients for CAD atoms in layer 1, 2 and 3 is around 9\%, 14\% and 22\%, respectively. This means the feature representation becomes less sparse with the increase of layers. A denser signal representation is helpful for modelling more complex signals which requires the use of more synthesis atoms for a good reconstruction quality. 

\subsection{Comparison with Deep Neural Networks}
\label{sec:compareDNNs}

To have a better understanding of DeepAM, in this section, we compare our proposed DeepAM method with deep neural networks (DNNs) learned using back-propagation algorithm. {We will first discuss the ability to learn from different training data size, in particular, small training data. We will then compare different methods on image super-resolution with noise to promote a better understanding of DeepAM.}

{For training DeepAM, we use the default settings discussed in Section \ref{sec:paramsetting}. Unless otherwise specified,} the number of IPAD atoms in each layer is set to be $K_{\text{LR}}$ which is the rank of the input LR data.

For comparison, DNNs are learned with the same training data using gradient descent with back-propagation. {Let us denote with DNN-R and DNN-S the DNN with ReLU (with bias terms) as non-linearity and soft-thresholding (with soft-thresholds) as non-linearity, respectively. Note that all the dictionaries, bias terms in DNN-R and the soft-thresholds in DNN-S are learnable.} The architecture of DNN-S is the same as our DeepAM. The implementation of DNNs is based on Pytorch with Adam optimizer \cite{kingma2014adam}, using batch size $1024$, initial learning rate $5\times10^{-3}$, and decay rate $0.1$. The parameter setting has been tuned to achieve the best performance. The parameters of the DNNs are initialized using the default method in Pytorch.

\subsubsection{Image Super-Resolution Learning from Small and Large Training Samples}

{In this part, we will discuss the ability of DeepAM to learn from training datasets of different sizes.}

{
Table \ref{tab:CompareDNN} shows the performance of DNN-R, DNN-S and DeepAM which are learned using a small and a large set of training samples, respectively. All the models are with 3 layers and each layer has 256 atoms (neurons). For the training of DNNs, when the number of training samples is large, the total number of epochs is set to 250 and the learning rate decay step is $50$, and when the number of training samples is small, the total number of epochs is set to 2500 and the learning rate decay step is $500$.}

{From the table, we can see that DNNs and DeepAM perform differently when learned from a small and a large number of training samples. In general, the performance of DeepAM is more stable compared to DNN-R and DNN-S.
When the number of training samples is sufficiently large, DNN-S is able to achieve better performances compared to DNN-R and DeepAM. The performance of DeepAM is around 0.2 dB and 0.1 dB lower than that of DNN-S on \textit{Set5} \cite{yang2010image} and \textit{Set14} \cite{zeyde2010single}, respectively.
When the number of training samples is relatively small, DeepAM achieves a higher PSNR compared to DNN-R and DNN-S. For example, when evaluated on \textit{Set5} \cite{yang2010image}, the performance of DeepAM is around 0.9 dB and 0.3 dB higher than that of DNN-R and DNN-S, respectively. This suggests that the proposed DeepAM learning method can be very useful for applications with limited number of training samples. 
}

{We further compare DeepAM with DNN-R and DNN-S on the self-example (or self-learning) image super-resolution  task \cite{bevilacqua2014single, shocher2018zero} which is an example of applications with limited number of training samples.}
In self-learning, these samples are obtained from the input low-resolution image and its down-sampled version since the high-resolution image is not available.
The learned image super-resolution model is then applied on the input low-resolution image to perform up-sampling.
In self-example image super-resolution, the number of training samples is usually very small.
For example, around $3200$ training sample pairs can be extracted from an input low-resolution image of resolution $128 \times 128$ pixels and its down-sampled version. As the number of training data is small, in order to achieve good performance, the total number of epochs for training DNNs is increased to 2500 with learning rate decay step 500. Similarly, we increase the number of iterations for training DeepAM to 5000.

{
Table \ref{tab:Set5_Self} shows the performance of DeepAM, DNN-S and DNN-R applied to self-example image super-resolution. 
From the table, we can see that DeepAM significantly outperforms DNN-S and DNN-R. The average PSNR of the proposed DeepAM is around 0.8 dB and 0.4 dB higher than that of DNN-R and DNN-S, respectively. In particular, for the image butterfly, the PSNR achieved by DeepAM is around 1.7 dB and 0.9 dB higher than that of DNN-R and DNN-S, respectively. The results further indicates that DeepAM can be very useful for applications with limited number of training samples.
}

\begin{table}[t]
    \centering
    \begin{tabular}{|c|C{1.3cm}|C{1.3cm}|C{1.3cm}|C{1.3cm}|}
    \hline 
    {Images} &  {Bicubic} &  {DNN-R} &  {DNN-S} &   {DeepAM} \tabularnewline
    \hline \hline 
baby     & 36.89 & 38.22 & 38.26 & \textbf{38.41} \\
bird    & 36.73 & 39.53 & 40.39 & \textbf{40.71} \\
butterfly     & 27.58 & 30.16 & 30.93 & \textbf{31.88} \\
head & 34.70 & 35.44 & {35.54} & \textbf{35.56} \\
woman      & 32.30 & 34.56 & 34.74 & \textbf{35.16} \\ \hline
Average    & 33.64 & 35.58 & 35.97 & \textbf{36.34} \\ \hline 
    \end{tabular}
    \caption{{PSNR (dB) of different methods for self-example image super-resolution evaluated on \textit{Set5} \cite{yang2010image}. (The best result in each row is in bold.)}}
    \label{tab:Set5_Self}
\end{table}

\begin{figure}[t]
    \centering
    \includegraphics[width=0.45\textwidth]{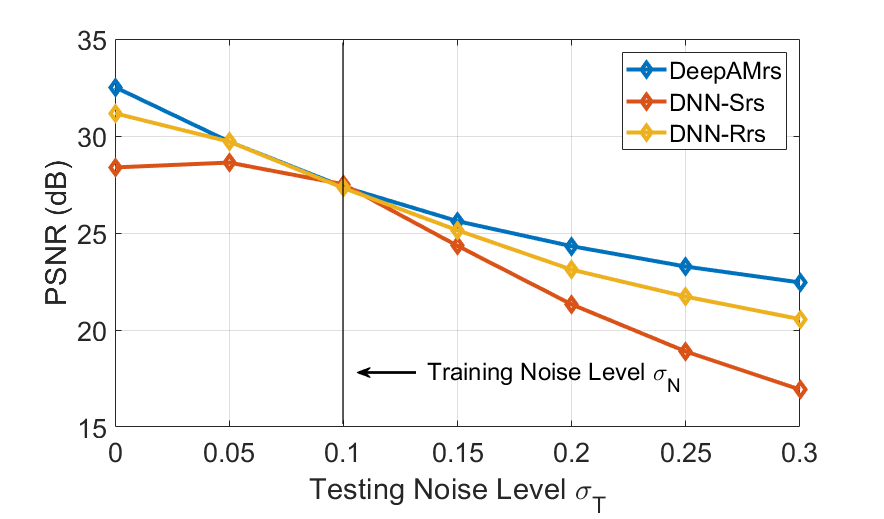}
    \caption{{The performance of $\text{DeepAM}_{\text{rs}}$, $\text{DNN-R}_{\text{rs}}$, and $\text{DNN-S}_{\text{rs}}$ under different levels of noise evaluated on \textit{Set5} \cite{yang2010image}. The training noise level is $\sigma_N = 0.1$, and the testing noise level $\sigma_T$ ranges from 0 to 0.3. The pixel values are within $[0,1]$.}}
    \label{fig:robustness}
\end{figure}

\begin{figure*}[hbt!]
	\centering
	\hspace*{\fill}
	\subfigure[Input LR image $\sigma_T=0$.]{
		\label{fig:Bird_LR_noise00} 
		\includegraphics[width=0.11\textwidth]{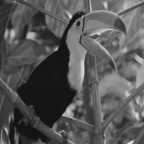}}
	\hfill
	\subfigure[$\text{DNN-R}_{\text{rs}}$ (PSNR = 33.34dB).]{
		\label{fig:Bird_LR_noise00_DNNRrs} 
		\includegraphics[width=0.22\textwidth]{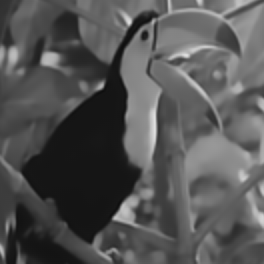}}
	\hfill
	\subfigure[$\text{DNN-S}_{\text{rs}}$ (PSNR = 29.01dB).]{
		\label{fig:Bird_LR_noise00_DNNSrs} 
		\includegraphics[width=0.22\textwidth]{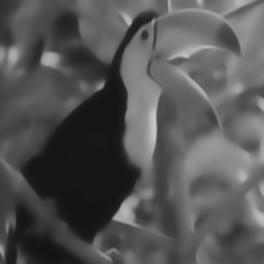}}
	\hfill
	\subfigure[$\text{DeepAM}_{\text{rs}}$ (PSNR = 35.55dB).]{
		\label{fig:Bird_LR_noise00_DeepAMrs} 
		\includegraphics[width=0.22\textwidth]{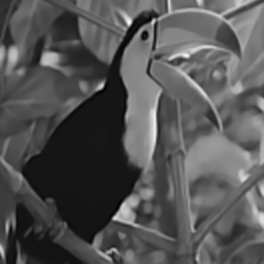}}
	\hspace*{\fill}
	
	\hspace*{\fill}
	\subfigure[Input LR image $\sigma_T=0.2$.]{
		\label{fig:Bird_LR_noise02} 
		\includegraphics[width=0.11\textwidth]{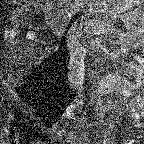}}
	\hfill
	\subfigure[$\text{DNN-R}_{\text{rs}}$ (PSNR = 23.90dB).]{
		\label{fig:Bird_LR_noise02_DNNRrs} 
		\includegraphics[width=0.22\textwidth]{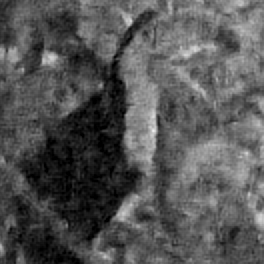}}
	\hfill
	\subfigure[$\text{DNN-S}_{\text{rs}}$ (PSNR = 21.65dB).]{
		\label{fig:Bird_LR_noise02_DNNSrs} 
		\includegraphics[width=0.22\textwidth]{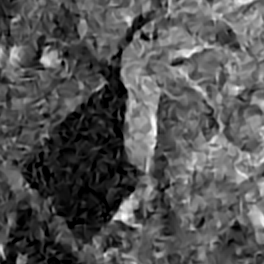}}
	\hfill
	\subfigure[$\text{DeepAM}_{\text{rs}}$ (PSNR = 25.07dB).]{
		\label{fig:Bird_LR_noise02_DeepAMrs} 
		\includegraphics[width=0.22\textwidth]{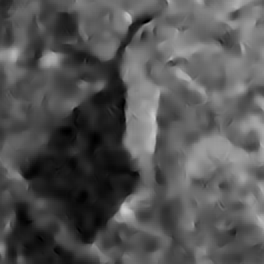}}
	\hspace*{\fill}
	
	\caption{{Image super-resolution with different noise levels using $\text{DNN-R}_{\text{rs}}$, $\text{DNN-S}_{\text{rs}}$ and $\text{DeepAM}_{\text{rs}}$. All the models are learned with training images with noise level $\sigma_N=0.1$. The thresholds (bias terms) are rescaled to adapt to the testing noise levels.}}
    \label{fig:SRnoise}
\end{figure*}

\subsubsection{Image Super-Resolution with Noise}

{In this part, we compare DeepAM with DNN-R and DNN-S on the noisy image super-resolution task. We will see that the learned DeepAM can be well adapted to testing images with unseen noise levels by rescaling the thresholds of IPAD and CAD at the first layer.}

{We use a noisy training dataset where noisy images are obtained by adding i.i.d. zero mean Gaussian noise $\mathcal{N}(0,\sigma_{N}^2)$ with $\sigma_N = 0.1$ and around 320,000 training patch pairs are used for training. As the input LR patches are noisy, we set the number of IPAD atoms in the first layer to $3 \times K_{\text{LR}}$ and the number of IPAD atoms in other layers to $K_{\text{LR}}$. 
DNNs are learned with weight decay $10^{-6}$ to achieve a more robust model.
}

{The test images are corrupted with i.i.d. zero mean Gaussian noise $\mathcal{N}(0,\sigma_{T}^2)$ and $\sigma_{T}$ might be different from $\sigma_{N}$. We assume that the testing noise levels is known since there are methods \cite{donoho1995noising, liu2006noise}  that can estimate the noise level from noisy images. When $\sigma_{T}^2 \neq \sigma_{N}^2$, we are faced with a case of mismatch between testing and training. However, since we know the task that each part of DeepAM has to fulfill, we can adjust the thresholds accordingly at testing stage without the need to retrain the network. This is something more difficult to achieve with a generic DNN.}

{From the analysis in Section \ref{sec:IPAD}, the IPAD thresholds should be proportional to the variance of the noise. Based on that, we can adapt the learned DeepAM to different noise levels by rescaling the IPAD thresholds of the first layer with a factor $\sigma_T^2 / \sigma_N^2$. The IPAD thresholds rescaling should be able to enhance the model robustness to a different noise level by passing essential information from the input signal.
With noisy input image, the function of CAD at the first layer and its thresholds can be interpreted as identifying meaningful signals from the noisy input. As the CAD and its thresholds are already more selective than IPAD part, we therefore rescale the CAD thresholds of the first layer with a factor $\sigma_T / \sigma_N$. We denote $\text{DeepAM}_{\text{rs}}$ as the DeepAM with rescaled IPAD and CAD thresholds. }

{
We can also apply bias terms (thresholds) rescaling to DNN-R and DNN-S to enhance the model robustness to unseen noise levels, and we denote with $\text{DNN-R}_{\text{rs}}$ and $\text{DNN-S}_{\text{rs}}$ as DNN-R and DNN-S with rescaled bias terms (thresholds) at the first layer with a factor $\sigma_T^2 / \sigma_N^2$, respectively. All the thresholds are rescaled with the same factor as we can not identify the function of the neurons in DNNs.
}

{
Fig. \ref{fig:robustness} shows the performance of $\text{DeepAM}_{\text{rs}}$, $\text{DNN-R}_{\text{rs}}$ and $\text{DNN-S}_{\text{rs}}$ evaluated with different noise levels.
The testing images are from \textit{Set5} \cite{yang2010image} and are with additive i.i.d. zero mean Gaussian noise $\mathcal{N}(0,\sigma_T^2)$ with $\sigma_T$ ranging from 0 to 0.3. 
From Fig. \ref{fig:robustness}, we can see that DeepAM achieves similar performance as DNN-R and DNN-S when evaluated on the testing noise level $\sigma_T = \sigma_N$. By rescaling IPAD thresholds and CAD thresholds, $\text{DeepAM}_{\text{rs}}$ achieves a robust adaptation of DeepAM to testing images with unseen noise levels.
The IPAD and its rescaled thresholds perform a denoising operation and pass essential information of the input signal to the next layer. 
The CAD and its rescaled thresholds shows a stronger ability to select meaningful signals from the input signals. The results of $\text{DeepAM}_{\text{rs}}$ shows the importance of understanding the function of each component in DeepAM.
Rescaling the bias terms (thresholds) of DNN-R and DNN-S can also improve their robustness to unseen noise levels, but cannot lead to the same performance as $\text{DeepAM}_{\text{rs}}$.} 

{Fig. \ref{fig:SRnoise} show two examples of image super-resolution results of $\text{DNN-R}_{\text{rs}}$, $\text{DNN-S}_{\text{rs}}$ and $\text{DeepAM}_{\text{rs}}$ on testing image \textit{bird} from \textit{Set5} \cite{zeyde2010single} with $\sigma_T = 0$ and $\sigma_T = 0.2$, respectively. 
$\text{DeepAM}_{\text{rs}}$ is able to achieve effective image super-resolution on both cases. The rescaled thresholds can successfully pass essential information and remove noise from the input signals. With rescaled thresholds, all methods can adapt to testing images with unseen noise levels. Taking the results with $\sigma_T = 0.2$ as an example, $\text{DNN-R}_{\text{rs}}$, $\text{DNN-S}_{\text{rs}}$ and $\text{DeepAM}_{\text{rs}}$ improves DNN-R, DNN-S and DeepAM by 4.06 dB, 1.64 dB and 5.05 dB, respectively.}

\begin{table*}[]
    \centering
   \begin{tabular}{|C{1.5cm}|C{1.5cm}|C{1.5cm}|C{1.5cm}|C{1.5cm}|C{1.5cm}|C{1.5cm}|C{1.5cm}|C{1.5cm}|}
    \hline 
    {Methods} &  {Bicubic} &  {SC \cite{zeyde2010single}} &  {ANR \cite{timofte2013anchored}} &  {A+ \cite{timofte2014a+}} &  {SRCNN \cite{dong2014learning}} &  {DNN-S} &  {DeepAM} &  {$\text{DeepAM}_{\text{bp}}$} \tabularnewline
    \hline \hline 
    \textit{Set5}               & 33.66   & 35.24 & 35.36 & 34.44 & 35.87 & 35.62 & 35.96  & \textbf{36.03} \\ \hline
    \textit{Set14}              & 30.23   & 31.50 & 31.49 & 31.26 & 31.88 & 31.63 & 31.87  & \textbf{31.92} \\ \hline 
    \end{tabular}
    \caption{{PSNR (dB) of different methods learned using a small training dataset evaluated on \textit{Set5} \cite{yang2010image} and \textit{Set14} \cite{zeyde2010single}. (The best result in each row is in bold.)}}
    \label{tab:smallTrainSet}
\end{table*}

\subsection{Comparison with Single Image Super-Resolution Methods}
\label{sec:compareSISR}

In this section, we will compare our proposed DeepAM method with some existing single image super-resolution methods including Bicubic interpolation, SC-based method \cite{zeyde2010single}, Anchored Neighbor Regression (ANR) method \cite{timofte2013anchored}, Adjusted Anchored Neighborhood Regression (A+) method \cite{timofte2014a+}, Super-Resolution Convolutional Neural Network (SRCNN) method \cite{dong2014learning}, {and DNN-S}. 

The SC-based method \cite{zeyde2010single} is a synthesis dictionary based method with a coupled LR and HR dictionary. The LR dictionary is learned using K-SVD \cite{aharon2006k} and has 1024 atoms, and the HR dictionary is learned using least squares. It assumes that a LR patch and its corresponding HR patch share the same sparse code which is retrieved using OMP \cite{pati1993orthogonal}. 
The input LR feature is the concatenation of the intensity, the first-order derivatives, and the second-order derivatives of the LR data and is further compressed using Principal Component Analysis (PCA). 
The ANR method \cite{timofte2013anchored} and the A+ method \cite{timofte2014a+} use the same feature representation as \cite{zeyde2010single}. They apply a learned LR synthesis dictionary for LR patch clustering and have a regression model for each dictionary atom. The super-resolution algorithm finds the nearest neighbor atom for each input LR signal and apply the corresponding regression model for HR signal prediction. The dictionary has 1024 atoms and thus there are 1024 regression models. The A+ method \cite{timofte2014a+} represented the state-of-the-art before the emergence of methods based on deep convolutional neural networks. 
The aforementioned methods \cite{zeyde2010single, timofte2013anchored, timofte2014a+} are all patch-based. 
The Super-Resolution Convolutional Neural Network (SRCNN) method \cite{dong2014learning, dong2015analysis} is the first to use convolutional neural network for single image super-resolution. SRCNN \cite{dong2014learning} has 3 layers and is with $64$ filters with spatial size $9 \times 9$, $32$ filters with spatial size $1 \times 1$ and $32$ filters with spatial size $5 \times 5$ for layer 1, 2 and 3, respectively. It takes the Bicubic up-scaled image as input and is able to upscale the input LR image without dividing the input image into patches. {SRCNN \cite{dong2015image} has achieved further improvement by using deeper and wider networks, and learning from ImageNet with 395,909 images.}
{$\text{DeepAM}_{\text{bp}}$ represents the 3-layer DeepAM (each layer with 256 atoms) that is refined using back-propagation. The input data for DeepAM and $\text{DeepAM}_{\text{bp}}$ is the intensity of the LR image patches.}

\begin{table*}[]
    \centering
   \begin{tabular}{|C{1.5cm}|C{1.5cm}|C{1.5cm}|C{1.5cm}|C{1.5cm}|C{1.5cm}|C{1.5cm}|C{1.5cm}|C{1.5cm}|}
    \hline 
    {Methods} &  {Bicubic} &  {SC \cite{zeyde2010single}} &  {ANR \cite{timofte2013anchored}} &  {A+ \cite{timofte2014a+}} &  {SRCNN \cite{dong2015image}} &  {DNN-S} &  {DeepAM} &  {$\text{DeepAM}_{\text{bp}}$} \tabularnewline
    \hline \hline 
    \textit{Set5}               & 33.66   & 35.78 & 35.82 & 36.23 & 36.66 & 36.62 & 36.44  & \textbf{36.71} \\ \hline
    \textit{Set14}              & 30.23   & 31.79 & 31.78 & 32.04 & \textbf{32.45} & 32.30 & 32.17  &  {32.40} \\ \hline
    \end{tabular}
    \caption{{PSNR (dB) of different methods learned using large training datasets evaluated on \textit{Set5} \cite{yang2010image} and \textit{Set14} \cite{zeyde2010single}. (The best result in each row is in bold.)}}
    \label{tab:largeTrainSet}
\end{table*}

In Table \ref{tab:smallTrainSet}, {we compare different methods in a small training dataset setting in which the patch-based methods \cite{zeyde2010single, timofte2013anchored, timofte2014a+} learn from 5000 patch pairs, and the SRCNN method \cite{dong2014learning} learns from 512 patch pairs of size $36 \times 36$ extracted from the 91 training images \cite{yang2010image}.}
{In this small training dataset setting, DeepAM already achieves better performance than the comparison methods and $\text{DeepAM}_{\text{bp}}$ further improves DeepAM and achieves the highest average PSNR.}
{
In Table \ref{tab:largeTrainSet}, we further compare different methods in a large training dataset setting. The patch-based methods \cite{zeyde2010single, timofte2013anchored, timofte2014a+} learn from 320000 patch pairs extracted from the 91 training images \cite{yang2010image}, and the SRCNN \cite{dong2015image} learns using more than 5 million patch pairs of size $33 \times 33$ from the ILSVRC 2013 ImageNet detection training partition.
In this large training dataset setting, the performance of DeepAM is around 0.2 dB lower than DNN-S. With further fine-tuning, $\text{DeepAM}_{\text{bp}}$ achieves the highest PSNR and its performance is comparable to that of SRCNN \cite{dong2015image}.
The fact that DeepAM and DNN perform similarly suggests that back-propagation optimization probably does something similar to our optimization strategy. Moreover, the good performance of $\text{DeepAM}_{\text{bp}}$ suggests that our approach provides an effective initialization for back-propagation.}

\section{Conclusions}

In this paper, we proposed a Deep Analysis Dictionary Model (DeepAM) which consists of multiple layers of analysis dictionary and soft-thresholding operators and a layer of synthesis dictionary. Each analysis dictionary has been designed to contain two sub-dictionaries: an Information Preserving Analysis Dictionary (IPAD) and a Clustering Analysis Dictionary (CAD). The IPAD and threshold pairs are to pass key information from the input to deeper layers. The function of the CAD and threshold pairs is to facilitate discrimination of key features. We proposed an extension of GOAL \cite{hawe2013analysis} to perform dictionary learning for both the IPAD and the CAD. The thresholds have been efficiently set according to simple principles, while leading to effective models. 
{Simulation results show that our proposed DeepAM outperforms DNNs with similar number of trainable parameters for small training datasets. 
We also show that, due to the interpretability of the components of DeepAM architecture, it is easier to adjust some parameters of the model to handle noisy test images with unseen levels of noise.
}

\ifCLASSOPTIONcaptionsoff
  \newpage
\fi

\bibliographystyle{IEEEtran}
\bibliography{bibs}

\begin{thebibliography}{10}
\providecommand{\url}[1]{#1}
\csname url@samestyle\endcsname
\providecommand{\newblock}{\relax}
\providecommand{\bibinfo}[2]{#2}
\providecommand{\BIBentrySTDinterwordspacing}{\spaceskip=0pt\relax}
\providecommand{\BIBentryALTinterwordstretchfactor}{4}
\providecommand{\BIBentryALTinterwordspacing}{\spaceskip=\fontdimen2\font plus
\BIBentryALTinterwordstretchfactor\fontdimen3\font minus
  \fontdimen4\font\relax}
\providecommand{\BIBforeignlanguage}[2]{{%
\expandafter\ifx\csname l@#1\endcsname\relax
\typeout{** WARNING: IEEEtran.bst: No hyphenation pattern has been}%
\typeout{** loaded for the language `#1'. Using the pattern for}%
\typeout{** the default language instead.}%
\else
\language=\csname l@#1\endcsname
\fi
#2}}
\providecommand{\BIBdecl}{\relax}
\BIBdecl

\bibitem{lecun2015deep}
Y.~LeCun, Y.~Bengio, and G.~Hinton, ``Deep learning,'' \emph{nature}, vol. 521,
  no. 7553, p. 436, 2015.

\bibitem{rumelhart1985learning}
D.~E. Rumelhart, G.~E. Hinton, and R.~J. Williams, ``Learning internal
  representations by error propagation,'' California Univ San Diego La Jolla
  Inst for Cognitive Science, Tech. Rep., 1985.

\bibitem{mallat2012group}
S.~Mallat, ``Group invariant scattering,'' \emph{Communications on Pure and
  Applied Mathematics}, vol.~65, no.~10, pp. 1331--1398, 2012.

\bibitem{bruna2013invariant}
J.~Bruna and S.~Mallat, ``Invariant scattering convolution networks,''
  \emph{IEEE Transactions on Pattern Analysis and Machine Intelligence},
  vol.~35, no.~8, pp. 1872--1886, 2013.

\bibitem{zeiler2014visualizing}
M.~D. Zeiler and R.~Fergus, ``Visualizing and understanding convolutional
  networks,'' in \emph{European conference on computer vision}.\hskip 1em plus
  0.5em minus 0.4em\relax Springer, 2014, pp. 818--833.

\bibitem{lei2018geometric}
N.~Lei, Z.~Luo, S.-T. Yau, and D.~X. Gu, ``Geometric understanding of deep
  learning,'' \emph{arXiv preprint arXiv:1805.10451}, 2018.

\bibitem{montufar2014number}
G.~F. Montufar, R.~Pascanu, K.~Cho, and Y.~Bengio, ``On the number of linear
  regions of deep neural networks,'' in \emph{Advances in neural information
  processing systems}, 2014, pp. 2924--2932.

\bibitem{giryes2016deep}
R.~Giryes, G.~Sapiro, and A.~M. Bronstein, ``Deep neural networks with random
  {G}aussian weights: A universal classification strategy?'' \emph{IEEE Trans.
  Signal Processing}, vol.~64, no.~13, pp. 3444--3457, 2016.

\bibitem{elad2010sparse}
M.~Elad, ``Sparse and redundant representations: From theory to applications in
  signal and image processing,'' \emph{Springer}, 2010.

\bibitem{elad2007analysis}
M.~Elad, P.~Milanfar, and R.~Rubinstein, ``Analysis versus synthesis in signal
  priors,'' \emph{Inverse problems}, vol.~23, no.~3, p. 947, 2007.

\bibitem{pati1993orthogonal}
Y.~C. Pati, R.~Rezaiifar, and P.~S. Krishnaprasad, ``Orthogonal matching
  pursuit: Recursive function approximation with applications to wavelet
  decomposition,'' in \emph{Proceedings of 27th Asilomar conference on signals,
  systems and computers}.\hskip 1em plus 0.5em minus 0.4em\relax IEEE, 1993,
  pp. 40--44.

\bibitem{dai2009subspace}
W.~Dai and O.~Milenkovic, ``Subspace pursuit for compressive sensing signal
  reconstruction,'' \emph{IEEE Transactions on Information Theory}, vol.~55,
  no.~5, pp. 2230--2249, 2009.

\bibitem{blumensath2009iterative}
T.~Blumensath and M.~E. Davies, ``Iterative hard thresholding for compressed
  sensing,'' \emph{Applied and computational harmonic analysis}, vol.~27,
  no.~3, pp. 265--274, 2009.

\bibitem{tibshirani1996regression}
R.~Tibshirani, ``Regression shrinkage and selection via the lasso,''
  \emph{Journal of the Royal Statistical Society. Series B (Methodological)},
  pp. 267--288, 1996.

\bibitem{chen2001atomic}
S.~S. Chen, D.~L. Donoho, and M.~A. Saunders, ``Atomic decomposition by basis
  pursuit,'' \emph{SIAM review}, vol.~43, no.~1, pp. 129--159, 2001.

\bibitem{daubechies2004iterative}
I.~Daubechies, M.~Defrise, and C.~De~Mol, ``An iterative thresholding algorithm
  for linear inverse problems with a sparsity constraint,''
  \emph{Communications on Pure and Applied Mathematics: A Journal Issued by the
  Courant Institute of Mathematical Sciences}, vol.~57, no.~11, pp. 1413--1457,
  2004.

\bibitem{beck2009fast}
A.~Beck and M.~Teboulle, ``A fast iterative shrinkage-thresholding algorithm
  for linear inverse problems,'' \emph{SIAM journal on imaging sciences},
  vol.~2, no.~1, pp. 183--202, 2009.

\bibitem{engan1999method}
K.~Engan, S.~O. Aase, and J.~H. Husoy, ``Method of optimal directions for frame
  design,'' in \emph{Acoustics, Speech, and Signal Processing, 1999.
  Proceedings., 1999 IEEE International Conference on}, vol.~5.\hskip 1em plus
  0.5em minus 0.4em\relax IEEE, 1999, pp. 2443--2446.

\bibitem{aharon2006k}
M.~Aharon, M.~Elad, A.~Bruckstein \emph{et~al.}, ``K-{SVD}: An algorithm for
  designing overcomplete dictionaries for sparse representation,'' \emph{IEEE
  Transactions on Signal Processing}, vol.~54, no.~11, p. 4311, 2006.

\bibitem{seghouane2018consistent}
A.-K. Seghouane and A.~Iqbal, ``Consistent adaptive sequential dictionary
  learning,'' \emph{Signal Processing}, vol. 153, pp. 300--310, 2018.

\bibitem{rubinstein2013analysis}
R.~Rubinstein, T.~Peleg, and M.~Elad, ``Analysis {K-SVD}: A dictionary-learning
  algorithm for the analysis sparse model,'' \emph{IEEE Transactions on Signal
  Processing}, vol.~61, no.~3, pp. 661--677, 2013.

\bibitem{yaghoobi2013constrained}
M.~Yaghoobi, S.~Nam, R.~Gribonval, and M.~E. Davies, ``Constrained overcomplete
  analysis operator learning for cosparse signal modelling,'' \emph{IEEE
  Transactions on Signal Processing}, vol.~61, no.~9, pp. 2341--2355, 2013.

\bibitem{dong2015analysis}
J.~Dong, W.~Wang, W.~Dai, M.~D. Plumbley, Z.-F. Han, and J.~Chambers,
  ``Analysis simco algorithms for sparse analysis model based dictionary
  learning,'' \emph{IEEE Transactions on Signal Processing}, vol.~64, no.~2,
  pp. 417--431, 2015.

\bibitem{ravishankar2013learning}
S.~Ravishankar and Y.~Bresler, ``Learning sparsifying transforms,'' \emph{IEEE
  Transactions on Signal Processing}, vol.~61, no.~5, pp. 1072--1086, 2012.

\bibitem{pfister2018learning}
L.~Pfister and Y.~Bresler, ``Learning filter bank sparsifying transforms,''
  \emph{IEEE Transactions on Signal Processing}, vol.~67, no.~2, pp. 504--519,
  2018.

\bibitem{hawe2013analysis}
S.~Hawe, M.~Kleinsteuber, and K.~Diepold, ``Analysis operator learning and its
  application to image reconstruction,'' \emph{IEEE Transactions on Image
  Processing}, vol.~22, no.~6, pp. 2138--2150, 2013.

\bibitem{kiechle2015bimodal}
M.~Kiechle, T.~Habigt, S.~Hawe, and M.~Kleinsteuber, ``A bimodal co-sparse
  analysis model for image processing,'' \emph{International Journal of
  Computer Vision}, vol. 114, no. 2-3, pp. 233--247, 2015.

\bibitem{rubinstein2014dictionary}
R.~Rubinstein and M.~Elad, ``Dictionary learning for analysis-synthesis
  thresholding,'' \emph{IEEE Transactions on Signal Processing}, vol.~62,
  no.~22, pp. 5962--5972, 2014.

\bibitem{rubinstein2010double}
R.~Rubinstein, M.~Zibulevsky, and M.~Elad, ``Double sparsity: Learning sparse
  dictionaries for sparse signal approximation,'' \emph{IEEE Transactions on
  Signal Processing}, vol.~58, no.~3, pp. 1553--1564, 2010.

\bibitem{seghouane2017basis}
A.-K. Seghouane and A.~Iqbal, ``Basis expansion approaches for regularized
  sequential dictionary learning algorithms with enforced sparsity for f{MRI}
  data analysis,'' \emph{IEEE Transactions on Medical Imaging}, vol.~36, no.~9,
  pp. 1796--1807, 2017.

\bibitem{papyan2017convolutional}
V.~Papyan, Y.~Romano, and M.~Elad, ``Convolutional neural networks analyzed via
  convolutional sparse coding,'' \emph{The Journal of Machine Learning
  Research}, vol.~18, no.~1, pp. 2887--2938, 2017.

\bibitem{sulam2017multi}
J.~Sulam, V.~Papyan, Y.~Romano, and M.~Elad, ``Multilayer convolutional sparse
  modeling: Pursuit and dictionary learning,'' \emph{IEEE Transactions on
  Signal Processing}, vol.~66, no.~15, pp. 4090--4104, 2018.

\bibitem{garcia2018convolutional}
C.~Garcia-Cardona and B.~Wohlberg, ``Convolutional dictionary learning: A
  comparative review and new algorithms,'' \emph{IEEE Transactions on
  Computational Imaging}, vol.~4, no.~3, pp. 366--381, 2018.

\bibitem{CAOL_Chun}
I.~Y. {Chun} and J.~A. {Fessler}, ``Convolutional analysis operator learning:
  Acceleration and convergence,'' \emph{IEEE Transactions on Image Processing},
  vol.~29, pp. 2108--2122, 2020.

\bibitem{CAOP_data_Chun}
I.~Y. {Chun}, D.~{Hong}, B.~{Adcock}, and J.~A. {Fessler}, ``Convolutional
  analysis operator learning: Dependence on training data,'' \emph{IEEE Signal
  Processing Letters}, vol.~26, no.~8, pp. 1137--1141, 2019.

\bibitem{tariyal2016deep}
S.~Tariyal, A.~Majumdar, R.~Singh, and M.~Vatsa, ``Deep dictionary learning,''
  \emph{IEEE Access}, vol.~4, pp. 10\,096--10\,109, 2016.

\bibitem{mahdizadehaghdam2018deep}
S.~Mahdizadehaghdam, A.~Panahi, H.~Krim, and L.~Dai, ``Deep dictionary
  learning: A parametric network approach,'' \emph{arXiv preprint
  arXiv:1803.04022}, 2018.

\bibitem{huang0418_DDSR}
J.-J. Huang and P.~L. Dragotti, ``A deep dictionary model for image
  super-resolution,'' in \emph{2018 IEEE International Conference on Acoustics,
  Speech and Signal Processing (ICASSP'18)}, March 2018.

\bibitem{huang0519_DDM_Key}
------, ``A deep dictionary model to preserve and disentangle key features in a
  signal,'' in \emph{2019 IEEE International Conference on Acoustics, Speech
  and Signal Processing (ICASSP'19)}, May 2019.

\bibitem{yang2010image}
J.~Yang, J.~Wright, T.~S. Huang, and Y.~Ma, ``Image super-resolution via sparse
  representation,'' \emph{IEEE Transactions on Image Processing}, vol.~19,
  no.~11, pp. 2861--2873, 2010.

\bibitem{zeyde2010single}
R.~Zeyde, M.~Elad, and M.~Protter, ``On single image scale-up using
  sparse-representations,'' in \emph{International conference on curves and
  surfaces}.\hskip 1em plus 0.5em minus 0.4em\relax Springer, 2010, pp.
  711--730.

\bibitem{timofte2013anchored}
R.~Timofte, V.~De~Smet, and L.~Van~Gool, ``Anchored neighborhood regression for
  fast example-based super-resolution,'' in \emph{Proceedings of the IEEE
  International Conference on Computer Vision}, 2013, pp. 1920--1927.

\bibitem{timofte2014a+}
------, ``A+: Adjusted anchored neighborhood regression for fast
  super-resolution,'' in \emph{Asian Conference on Computer Vision}.\hskip 1em
  plus 0.5em minus 0.4em\relax Springer, 2014, pp. 111--126.

\bibitem{huang2015fast}
J.-J. Huang, W.-C. Siu, and T.-R. Liu, ``Fast image interpolation via random
  forests,'' \emph{IEEE Transactions on Image Processing}, vol.~24, no.~10, pp.
  3232--3245, 2015.

\bibitem{huang1215_learning}
J.-J. Huang and W.-C. Siu, ``Learning hierarchical decision trees for single
  image super-resolution,'' \emph{IEEE Transactions on Circuits and Systems for
  Video Technology}, 2017.

\bibitem{huang0717_SRHRF+}
J.-J. Huang, T.~Liu, P.~L. Dragotti, and T.~Stathaki, ``{SRHRF+}: Self-example
  enhanced single image super-resolution using hierarchical random forests,''
  in \emph{IEEE Conference on Computer Vision and Pattern Recognition (CVPR)
  Workshop on New Trends in Image Restoration and Enhancement}, July 2017.

\bibitem{dong2014learning}
C.~Dong, C.~C. Loy, K.~He, and X.~Tang, ``Learning a deep convolutional network
  for image super-resolution,'' in \emph{European Conference on Computer
  Vision}.\hskip 1em plus 0.5em minus 0.4em\relax Springer, 2014, pp. 184--199.

\bibitem{dong2016image}
------, ``Image super-resolution using deep convolutional networks,''
  \emph{IEEE Transactions on Pattern Analysis and Machine Intelligence},
  vol.~38, no.~2, pp. 295--307, 2016.

\bibitem{kim2016accurate}
J.~Kim, J.~Kwon~Lee, and K.~Mu~Lee, ``Accurate image super-resolution using
  very deep convolutional networks,'' in \emph{Proceedings of the IEEE
  Conference on Computer Vision and Pattern Recognition}, 2016, pp. 1646--1654.

\bibitem{fawzi2015dictionary}
A.~Fawzi, M.~Davies, and P.~Frossard, ``Dictionary learning for fast
  classification based on soft-thresholding,'' \emph{International Journal of
  Computer Vision}, vol. 114, no. 2-3, pp. 306--321, 2015.

\bibitem{chun2018deep}
I.~Y. Chun and J.~A. Fessler, ``Deep {BCD}-net using identical
  encoding-decoding {CNN} structures for iterative image recovery,'' in
  \emph{2018 IEEE 13th Image, Video, and Multidimensional Signal Processing
  Workshop (IVMSP)}.\hskip 1em plus 0.5em minus 0.4em\relax IEEE, 2018.

\bibitem{chun2019bcd}
I.~Y. Chun, X.~Zheng, Y.~Long, and J.~A. Fessler, ``{BCD}-net for low-dose {CT}
  reconstruction: Acceleration, convergence, and generalization,'' in
  \emph{International Conference on Medical Image Computing and
  Computer-Assisted Intervention}.\hskip 1em plus 0.5em minus 0.4em\relax
  Springer, 2019.

\bibitem{chun2020momentum}
I.~Y. Chun, Z.~Huang, H.~Lim, and J.~Fessler, ``Momentum-net: Fast and
  convergent iterative neural network for inverse problems,'' \emph{IEEE
  Transactions on Pattern Analysis and Machine Intelligence}, 2020.

\bibitem{absil2009optimization}
P.-A. Absil, R.~Mahony, and R.~Sepulchre, \emph{Optimization algorithms on
  matrix manifolds}.\hskip 1em plus 0.5em minus 0.4em\relax Princeton
  University Press, 2009.

\bibitem{elad2006simple}
M.~Elad, ``Why simple shrinkage is still relevant for redundant
  representations?'' \emph{IEEE Transactions on Information Theory}, vol.~52,
  no.~12, pp. 5559--5569, 2006.

\bibitem{raphan2008optimal}
M.~Raphan and E.~P. Simoncelli, ``Optimal denoising in redundant
  representations,'' \emph{IEEE Transactions on Image Processing}, vol.~17,
  no.~8, pp. 1342--1352, 2008.

\bibitem{lin2006bayesian}
Y.~Lin and D.~D. Lee, ``Bayesian l 1-norm sparse learning,'' in
  \emph{Acoustics, Speech and Signal Processing, 2006. ICASSP 2006 Proceedings.
  2006 IEEE International Conference on}, vol.~5.\hskip 1em plus 0.5em minus
  0.4em\relax IEEE, 2006, pp. V--V.

\bibitem{kingma2014adam}
D.~P. Kingma and J.~Ba, ``Adam: A method for stochastic optimization,''
  \emph{arXiv preprint arXiv:1412.6980}, 2014.

\bibitem{bevilacqua2014single}
M.~Bevilacqua, A.~Roumy, C.~Guillemot, and M.-L.~A. Morel, ``Single-image
  super-resolution via linear mapping of interpolated self-examples,''
  \emph{IEEE Transactions on Image Processing}, vol.~23, no.~12, pp.
  5334--5347, 2014.

\bibitem{shocher2018zero}
A.~Shocher, N.~Cohen, and M.~Irani, ``“zero-shot” super-resolution using
  deep internal learning,'' in \emph{Proceedings of the IEEE conference on
  computer vision and pattern recognition}, 2018, pp. 3118--3126.

\bibitem{donoho1995noising}
D.~L. Donoho, ``De-noising by soft-thresholding,'' \emph{IEEE Transactions on
  Information Theory}, vol.~41, no.~3, pp. 613--627, 1995.

\bibitem{liu2006noise}
C.~Liu, W.~T. Freeman, R.~Szeliski, and S.~B. Kang, ``Noise estimation from a
  single image,'' in \emph{2006 IEEE Computer Society Conference on Computer
  Vision and Pattern Recognition (CVPR'06)}, vol.~1.\hskip 1em plus 0.5em minus
  0.4em\relax IEEE, 2006, pp. 901--908.

\bibitem{dong2015image}
C.~Dong, C.~C. Loy, K.~He, and X.~Tang, ``Image super-resolution using deep
  convolutional networks,'' \emph{IEEE Transactions on Pattern Analysis and
  Machine Intelligence}, vol.~38, no.~2, pp. 295--307, 2015.

\end{thebibliography}

\end{document}